\documentclass[preprint,12pt]{elsarticle}
\usepackage[a4paper, margin=0.5in]{geometry}
\usepackage{setspace}

\usepackage{amssymb}
\usepackage{amsmath}
\usepackage{natbib}  % Required for \citet and \citep
\usepackage{graphicx}%
\usepackage{multirow}%
\usepackage{amsmath,amssymb,amsfonts}%
\usepackage{amsthm}%
\usepackage{mathrsfs}%
\usepackage[title]{appendix}%
\usepackage{xcolor}%
\usepackage{textcomp}%
\usepackage{manyfoot}%
\usepackage{booktabs}%
\usepackage{algpseudocode}%
\usepackage{listings}%
\usepackage{subcaption}
\usepackage{hyperref}

\usepackage[ruled,vlined]{algorithm2e}
\usepackage{braket}
\usepackage{xcolor}
\usepackage{makecell}
\definecolor{highlight}{RGB}{133,223,255} 
\usepackage[dvipsnames,table]{xcolor}
\usepackage{quantikz}
\usepackage{tikz}

\journal{Nuclear Physics B}

\begin{document}

\begin{frontmatter}


% use the tnoteref command within \title for footnotes;
% use the tnotetext command for theassociated footnote;
% use the fnref command within \author or \affiliation for footnotes;
% use the fntext command for theassociated footnote;
% use the corref command within \author for corresponding author footnotes;
% use the cortext command for theassociated footnote;
% \author{Name\corref{cor1}\fnref{label2}}
% \affiliation{organization={},

\title{Hybrid Quantum-Classical Ridgelet Neural Network for Portfolio Optimization
}   %% Article title
 

\author{Bahadur Yadav, Sanjay Kumar Mohanty } %% Author name

% Author affiliation
\affiliation{organization={ Department of Mathematics, School of Advanced Sciences, Vellore Institute of Technology },%Department and Organization
           city={Vellore},
           postcode={632014}, 
            state={Tamil Nadu},
            country={India}}

%% Abstract
\newpage
\begin{abstract}
In this study, we introduce a quantum computing method that incorporates Ridglet transforms into quantum processing pipelines for financial time-series forecasting with Quantum Approximate Optimization Algorithm (QAOA)-based portfolio optimization. We propose a Quantum Ridgelet Neural Network (QRNN) model for forecasting time-series data that integrates Parametrized Quantum Circuits (PQCs) with ridgelet-based feature transformations and QAOA-based portfolio optimization for asset selection. By breaking down financial time-series data into multi-resolution components, the ridgelet transform enables the identification of both local and global trends. Ridgelet-based features improve the scalability and accuracy of quantum computing by significantly reducing the number of qubits needed. However, the predicted results are turned into a QUBO-based mean-variance optimization problem and solved with QAOA to select the best stocks. Our study begins with a theoretical formulation of the single-qubit system for our proposed model. This formulation is further extended to a multi-qubit system, and we show that it captures a significant fraction of the predictive signal. To investigate the practical feasibility of our proposed model under realistic circumstances, we further analyze three quantum configurations built upon the same PQC: (i) a noisy quantum model incorporating bit-flip, phase-flip, and depolarizing channels, (ii) a Quantum Error Correction (QEC) model based on a three-qubit repetition, and (iii) a partial QEC model that maintains a balance between error mitigation and computational efficiency. The performance and scalability of our model are evaluated by extensive experiments in single-qubit, two-qubit, and multi-qubit scenarios. Our findings show that while noise considerably reduces performance, adding QEC significantly enhances stability, with partial QEC giving a useful trade-off between accuracy and circuit complexity. Furthermore, our model is also evaluated against various quantum and classical deep learning models using real financial time-series data. These empirical findings show that our model achieves superior predictive performance, emphasizing its potential use in real-world financial forecasting and portfolio optimization tasks. For computational efficiency and scalability in higher-dimensional situations, we employed advanced NVIDIA H100 NVL GPU clusters.

\end{abstract}


% \begin{highlights}
% \end{highlights}

%% Keywords

\end{frontmatter}



\section{Introduction}
{I}n quantitative finance, portfolio optimization is a fundamental problem that aims to determine the optimal subset of assets for investment while balancing expected return and portfolio risk. However, in real-world financial markets, investors are often required to select a limited number of assets from a large universe of potential stocks to maximize their portfolio performance under predefined investment constraints. However, forecasting the financial market is also a highly lucrative area of research, offering promising opportunities for both researchers and investors. However, accurately forecasting financial markets is challenging due to the complex, non-linear, and chaotic nature, which necessitates the use of advanced and powerful methods. Fundamentally, prediction is the process of predicting future events by examining past evidence, which involves examining historical market activity to predict future price fluctuations in the context of stock prices. Many factors influence stock prices, including global economic trends, political situations, and the performance of individual companies \cite{adebiyi2012stock}. There are two main sources of information used in stock price prediction analysis: unstructured textual sources such as financial news and reports, and organized periodic past market data \cite{long2024hybrid}. In this context, a time series is a temporal sequence of observations for a particular variable that forms the basis of forecasting, such as daily stock prices, trade volumes, and various market factors.

Traditional statistical techniques are used in a wide range of research to forecast stock prices and optimize portfolios \cite{yang2015robust, abraham1983statistical}. In particular, stock price prediction is often performed using the Autoregressive Integrated Moving Average (ARIMA) and Generalized Autoregressive Conditional Heteroscedasticity (GARCH) models \cite{mutinda2024stock, kim2018forecasting}. Nevertheless, these statistical techniques are based on particular assumptions and might not adequately account for the significant non-linearity and volatility present in financial markets \cite{babu2015prediction}. To overcome these restrictions, hybrid strategies have been developed that integrate machine learning and deep learning methods, such as Long Short-Term Memory (LSTM) and XGBoost, with conventional statistical models to improve prediction reliability and precision \cite{wang2024novel, mehtarizadeh2025stock}.

Recently, Machine Learning (ML), particularly Deep Learning (DL), has revolutionized stock price prediction and portfolio optimization by enabling the identification of intricate models that capture complex, non-linear patterns in financial time series data \cite{lim2021time,ma2021portfolio}. Despite being fundamental, traditional statistical models such as GARCH and ARIMA often fail to account for the intrinsic instability and irregularities of stock markets \cite{bao2025data}. On the other hand, DL models that have shown greater accuracy in predicting stock prices, volatility, and trends include Transformer-based frameworks, LSTM networks, and Convolutional Neural Networks (CNNs) \cite{zhang2022transformer}.  These models improve predicted accuracy by utilizing immense amounts of real-time and past data, such as sentiment analysis and market signals \cite{pang2020innovative}. Nonetheless, issues persist with data quality, model interpretability, and the need for real-time flexibility \cite{zhang2018novel}. Resolving these issues is crucial for enhancing the practical application of ML and DL methods in financial forecasting.
 
Over the past few years, wavelet transformations and neural networks have gained increasing popularity among researchers for enhancing the predictive accuracy of financial time series data and portfolio management \cite{hsieh2011forecasting, yadav2026hybrid}. Initial studies demonstrated that wavelet decomposition could efficiently separate high- and low-frequency components in market statistics, enabling neural networks to model nonlinear temporal patterns \cite{vogl2022forecasting} more accurately. Further investigations revealed that wavelet-based neural networks significantly improved feature extraction and prediction performance on financial datasets \cite{chang2008hybrid}.
Despite these developments, early wavelet-neural methods often suffered from overfitting and sensitivity to wavelet parameter choices, making them less resilient under unstable market conditions. To address these obstacles, hybrid architectures that combine enhanced function-approximation techniques with kernel-based methods have been developed, providing greater prediction stability and generalization \cite{kilicc2023hybrid}. At the same time, functional wavelet–kernel methods showed improved learning performance for high-dimensional time series data \cite{antoniadis2006functional}.

To effectively manage the uncertainty and nonlinearity present in financial dynamics, additional improvements included fuzzy analysis and higher-order modeling techniques \cite{yang2018time}. However, because wavelets alone were unable to capture sudden directional changes in financial markets, researchers introduced ridgelet transforms, which improved prediction during volatile markets by providing sensitivity to singularities and directional information \cite{amjady2011short}. These developments were then expanded upon by models that incorporated ridgelet and wavelet transforms, and further improved using metaheuristic methods to reduce computational costs and increase efficiency \cite{leng2018new}. Furthermore, to minimize prediction error and overfitting across various financial market indices, wavelet and ridgelet-inspired models were combined with feature selection algorithms and ensemble learning techniques \cite{lei2018wavelet}. In adaptation to unpredictable market conditions, these ensemble-enhanced hybrid frameworks show greater resilience and flexibility \cite{singh2025wavelet}.
This development shows how hybrid ridgelet-enhanced and ensemble-based algorithms have replaced conventional wavelet-neural networks, each of which has addressed earlier drawbacks and helped create stock market forecasting techniques that are more precise, reliable, and comprehensible.

Recent years have seen tremendous advancements in quantum computing, notably the development of quantum algorithms and hardware \cite{wendin2017quantum, arute2019quantum, cong2019quantum}, which are closely correlated with advancements in deep learning. Large integer factorization is one of the computationally unsolvable issues that quantum computers could solve \cite{shor1999polynomial}. Based on this framework, Quantum Machine Learning (QML) has emerged as a promising multidisciplinary field that combines concepts from machine learning and quantum computation to enhance descriptive ability and learning efficiency. Earlier attempts in this field investigated quantum-inspired neural network models that utilized classical representations, notably the Bloch Sphere, to simulate quantum computation and predict stock prices \cite{azevedo2007time}. Nevertheless, neither qubits nor true quantum operations were used in the computation of these models. The creation of hybrid quantum-classical models that can learn complicated networks by optimizing tunable quantum variables has been made possible more recently by the discovery of variational quantum circuits (VQCs) \cite{mitarai2018quantum}. VQCs are more expressive than classical models due to the quantum entanglement between qubits, making them ideal for data-driven financial prediction problems, such as time-series forecasting.

Based on the previously mentioned developments, our study presents an integrated quantum-classical approach for financial forecasting and portfolio optimization. Our proposed approach combines a QRNN for multi-asset return forecasting with a QAOA-based portfolio optimization module for stock selection. The forecasting component uses ridgelet-based feature transformation and VQC processing to estimate the expected future returns of candidate assets. These predicted returns are then incorporated into a cardinality-constrained mean–variance portfolio optimization framework, reformulated as a QUBO problem, and solved using QAOA to identify the optimal subset of assets under return–risk trade-off constraints. Additionally, quantum noise and error effects are analyzed to evaluate the robustness and practical feasibility of our proposed framework under realistic quantum computing conditions.

The major contributions of our work are summarized as follows:
\begin{itemize}
    \item We propose a novel quantum ridgelet neural network model for forecasting the time series data that incorporates ridgelet-based feature transformation with PQCs, enabling efficient representation learning in quantum computing. However, ridgelet features enable our model to efficiently capture directional patterns and complex features in the data.
    \item We incorporate QRNN forecasting into a QAOA-based portfolio optimization framework and reformulate the portfolio section problem into QUBO form for quantum optimization.

\item Our proposed QRNN model is systematically formulated theoretically, beginning with a single-qubit system and sequentially expanding to a multi-qubit scenario, providing a robust mathematical basis for our framework.

\item Based on a corresponding PQCs, we construct and examine three quantum noise configurations: a noisy quantum model with a realistic noise configuration, a full QEC model with a three-qubit repetition algorithm, and a partial QEC model.

\item We conducted comprehensive experimentation in single-qubit, two-qubit, and multi-qubit systems for evaluating QEC techniques' efficacy, scalability, and performance degradability under noise.

\item We demonstrate superior prediction capabilities by comparing the performance of our proposed QRNN model with various quantum and classical deep learning models and validating it on actual financial time-series data.
\item The effectiveness of our model is further employed within a QAOA-based portfolio optimization strategy to identify high-potential stocks for portfolio creation.
\end{itemize}

This article is organized as follows. Section \ref{section:2}
 presents the fundamentals of quantum computing. Section \ref{section:3} presents the problem formulation of our proposed framework. Section \ref{section:4} describes the QRNN forecasting model. Section \ref{section:5} provides the details of the QAOA-based portfolio optimization framework. Section \ref{section:6} analyzes the quantum noise and the error effect in our model. Section \ref{section:7} provides the experimental result and discussion. Conclusion and future discussion are provided in 
section \ref{section:8}.

\section{Fundamentals of Quantum Computing}\label{section:2}
Here, we will discuss the fundamentals of quantum gates and the single quantum bit (Qubit) required for the construction of quantum ridgelet neural networks \cite{jacquier2022quantum, nielsen2010quantum}.
\subsection{Qubit and quantum gates}
In classical computing, information is represented using 
bits that take values in the set $\{0,1\}$. On the other hand, quantum information is 
encoded in Qubits, which are located in a two-dimensional 
complex Hilbert space $\mathcal{H} \cong \mathbb{C}^2$ with computational basis 
$\{\ket{0},\ket{1}\}$. The expression for a universal pure qubit state is
\(
\ket{\Phi} = \Phi_0\ket{0} + \Phi_1\ket{1},
\) where $\Phi_0, \Phi_1 \in \mathbb{C}$ are such that $|\Phi_0|^{2} + |\Phi_1|^{2} = 1$. 
Here, $|\Phi_0|^{2}$ and $|\Phi_1|^{2}$ refer the probabilities of the qubit 
taking the value $\ket{0} := \begin{pmatrix} 1 \\ 0 \end{pmatrix}$ and 
$\ket{1} := \begin{pmatrix} 0 \\ 1 \end{pmatrix}$, respectively. However, when measuring quantum information in qubits, a system is said to have $n $ qubits if its Hilbert space has $2^n$ dimensions.  Two orthogonal states of a single-qubit are written as $\ket{0},\; \ket{1}$. In a broader sense, $2n$ mutually orthogonal states of $n $ qubits can be expressed as $\{\,|i\rangle\,\}$, where $i$ is a binary number of $n $ bits.

Classical gates employ Boolean logic for controlling bits in classical computing. The most often used classical gates are XOR, AND, OR, and NOT. Transistors are used to physically implement these gates, which serve as the basis for classical digital computing. Whereas the basic procedures needed to control qubits in a quantum computer are known as quantum gates. Quantum gates are defined by unitary matrices.  Some of the quantum gates used in our work. The gates provided below are
$I $= $\begin{pmatrix} 1 & 0 \\ 0 & 1 \end{pmatrix}$, $ X$= $\begin{pmatrix} 0 & 1 \\ 1 & 0 \end{pmatrix}$,  $ Y$= $\begin{pmatrix} 0 & $-i$ \\ $i$ & 0 \end{pmatrix}$, $Z$= $\begin{pmatrix} 1 & 0 \\ 0 & -1 \end{pmatrix}$

These are some rotation gates that are used and  defined as follows: 
\begin{align*}    
R_x(\theta)&=\begin{pmatrix}\cos\frac{\theta}{2}&-i\sin\frac{\theta}{2}\\ -i\sin\frac{\theta}{2}&\cos\frac{\theta}{2}\end{pmatrix},
\
R_y(\theta)=\begin{pmatrix}\cos\frac{\theta}{2}&-\sin\frac{\theta}{2}\\ \ \ \sin\frac{\theta}{2}&\cos\frac{\theta}{2}\end{pmatrix}, 
\\ R_z(\theta)&=\begin{pmatrix}e^{-i\frac{\theta}{2}}&0\\0&e^{i\frac{\theta}{2}}\end{pmatrix}. 
\end{align*}
Here $i := \sqrt{-1}$ is an imaginary number $\theta \in \mathbb{R}$ is a parameter . Increasing the number of qubits is a difficult challenge for both improving accuracy and computing. As a result, we built a fundamental model using a single qubit in our framework.   

\subsection{Single-qubit system} 

\begin{figure*}[!ht]
    \centering
    % --------- Second figure ---------
    \includegraphics[width=12cm]{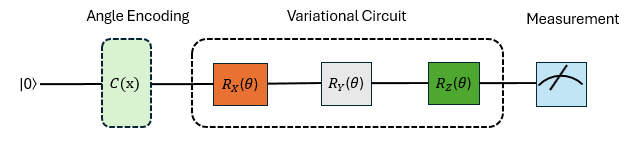}
    \caption{ The general architecture for a single-qubit VQC with angle encoding for embedding classical input data into the quantum state, followed by a parameterized rotation. The final layer is a measurement layer employed to measure the output.}
     \label{single qubit}
 \end{figure*}

In this subsection, we employ these quantum fundamentals to construct our model, based on the qubit formalism and unitary gate operations presented in the preceding section.
A two-dimensional complex Hilbert space is denoted by \(\mathcal{H} = \mathbb{C}^{2} \) and describes a single-qubit. We consider squared magnitudes of amplitudes as probabilities because this space has an inner product. We consider the standard computational basis, written as: \[ |0\rangle = \begin{pmatrix} 1\\[1mm] 0 \end{pmatrix}, \qquad |1\rangle = \begin{pmatrix} 0\\[1mm] 1 \end{pmatrix}. \]
These basis vectors physically describe the two distinct classical outcomes of measuring a qubit in the computational basis. A normalized superposition is a universal pure state of the qubit that can be expressed as:
\[ |\Phi\rangle = \Phi_{0}|0\rangle + \Phi_{1}|1\rangle, \qquad |\Phi_{0}|^{2} + |\Phi_{1}|^{2} = 1, \] 

where the complex coefficients $\Phi_{0}$ and $\Phi_{1}$ are constituted as  probability amplitudes. The probabilities measurement for the results $|0\rangle$ and $|1\rangle$, respectively, are obtained from their squared magnitudes. Thus, instead of deterministic values, quantum information is contained in the amplitudes.
A unitary operator $U \in U(2)$ that satisfies \(U^{\dagger} U = I_{2} \) represents the time evolution operations on a qubit.
Unitarity provides the conservation of total probability during the evolution by preserving the norm of the state vector. We concentrate on the following particular family of unitary gates in our framework:  rotations of the Bloch sphere around its $y$--axis. This is defined as 
\[ R_{y}(\theta) = e^{-i\theta\sigma_{y}/2} =\begin{pmatrix}\cos\frac{\theta}{2}&-\sin\frac{\theta}{2}\\ \  \ \sin\frac{\theta}{2}&\cos\frac{\theta}{2}\end{pmatrix},\]
where $\sigma_{y}$ represents the Pauli $Y$ matrix.  The Bloch sphere's state vector is geometrically rotated by $R_{y}(\theta)$ around the $y$--axis by an angle $\theta$. In the following section, we incorporate classical ridgelet neural networks into the quantum context; this rotation will operate as a nonlinear activation. A single-qubit framework is depicted in Fig. \ref{single qubit}.

\section{Problem Formulation of the Proposed Framework}\label{section:3}

In this section, we present a mathematical formulation of our proposed model, which is an integrated quantum-enhanced financial decision-making framework that effectively tackles financial time-series forecasting and portfolio optimization. Here, our proposed approach integrates a QRNN for predictive return estimation with a QAOA-based portfolio optimization module for optimal asset selection.

Let $X\in\mathbb{R}^{T\times N}$ represent the historical financial time-series data of $N$  assets over $T$ time periods. The first objective of our proposed framework is to learn a forecasting function, which is expressed as follows:

\begin{equation*}
\hat{\mu}=f_{\text{QRNN}}(X;\Theta),
\end{equation*}

where $f_{\text{QRNN}}(\cdot)$ represents our proposed QRNN parameterized by trainable parameters $\Theta$, and $\hat{\mu}\in\mathbb{R}^{N}$ represents the expected future return predicted for all vector assets. 

\begin{figure*}[!ht]
    \centering
    % --------- Second figure ---------
    \includegraphics[width=14cm]{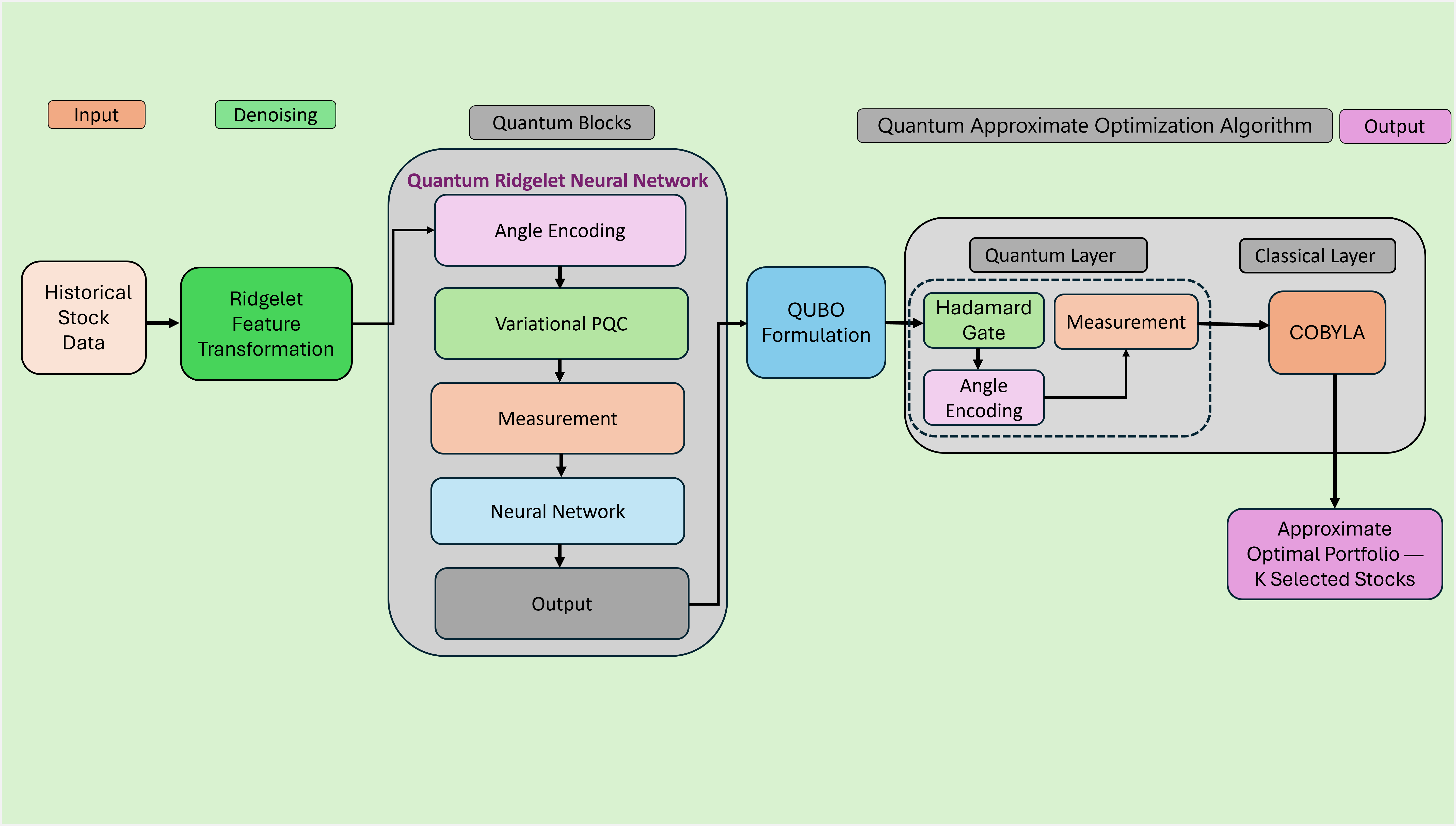}
    \caption{Systematic flow diagram of our proposed work.}
     \label{fig:QAOA}
 \end{figure*}

From the predicted return of the asset $\hat{\mu}$, now the second objective is to identify an optimal subset of assets for portfolio construction. Next, we define the binary portfolio selection vector $x\in\{0,1\}^{N}$, which is given as follows:

\begin{equation*}
x_i=
\begin{cases}
1, & \text{if asset } i \text{ is selected in the portfolio},\\
0, & \text{otherwise},
\end{cases}
\end{equation*}

for $i=1,2,\dots,N$. Furthermore, let $\Sigma\in\mathbb{R}^{N\times N}$ be the covariance matrix of historical asset returns, representing pairwise risk interactions among candidate assets. The portfolio optimization objective is formulated using the following cardinality-constrained mean-variance optimization problem, and is given as:

\begin{equation*}
\min_{x\in\{0,1\}^{N}}
-\hat{\mu}^{\top}x+x^{\top}\Sigma x
\end{equation*}

subject to

\begin{equation*}
\sum_{i=1}^{N}x_i=K,
\end{equation*}

where $K$ is the predefined portfolio cardinality specifying the number of assets to be selected.
Furthermore, to efficiently solve the above combinatorial optimization problem, the constrained binary optimization problem is transformed into a QUBO formulation and solved using the QAOA.
Fig. \ref{fig:QAOA} displays the overall workflow of the proposed integrated forecasting and optimization framework.
The following sections provide detailed descriptions of the QAOA-based portfolio optimization process and the QRNN predicting framework.

\section{Quantum Ridgelet Neural Network }\label{section:4}
In this section, we will discuss the QRNN approach. We first present a mathematical formulation of a single qubit system, subsequently extending the model to a multi-qubit system to capture the high-dimensional feature representation.
\subsection{QRNN for single-qubit system}
In this subsection, we present the mathematical details of the QRNN for a single-qubit system and discuss the integration of the Ridgelet transform with various types of quantum gates.

Let $\eta: \mathbb{R}^n \to \mathbb{R}$ be a Ridgelet transform defined as $\eta_{a,b, u}(x) = a^{-1/2} \eta\left(\frac{x \cdot\ u-b}{a}\right), \quad$\(
(u, a, b) \in \mathbb{R}^n \times \mathbb{R} \times \mathbb{R},
\)$
$  
where we refer to \( (a, b)\) as a parameter and  \( u\) is a vector of unit length. We define a function $g_{_J}:\mathbb{R}^{n}\to\mathbb{R}$, a finite ridgelet expansion is written as
\[ g_{_J}(x) = \sum_{i=1}^{J} c_{i}\, \eta_{u_{i},a_{i},b_{i}}(x) = \sum_{i=1}^{J} c_{i}\, a_{i}^{-1/2}\, \eta\!\left(\frac{x \cdot u_{i}-b_{i}}{a_{i}}\right), \] 
where $ J$ is a number of neurons, $ c_{i}$  as an output parameter, and $(a_{i}, b_{i})$ are trainable parameters. This expression serves as the foundation for a Classical Ridgelet Neural Network (CRNN), in which each term acts as a neuron with constraints $(a_{i}, b_{i})$ and  $\eta$ as an activation function. Furthermore, we employ an angle-encoding approach to integrate classical input data $x \in \mathbb{R}$ into a quantum state. Furthermore, we define  encoding function $\Psi:\mathbb{R} \to \mathbb{R}$ and the input state is defined as follows:
 \[ |\eta_{\mathrm{in}}(x)\rangle := R_{y}(\Psi(x))\,|0\rangle. \]
 Using the explicit matrix form of $R_{y}$, we get 
 \(\ |\eta_{\mathrm{in}}(x)\rangle = \cos\! \frac{\Psi(x)}{2}\  \!|0\rangle + \sin\!\frac{\Psi(x)}{2} \ \!|1\rangle. \)
 Thus, the classical value $x$ is encoded into the amplitudes of the qubit via a rotation angle $\Psi(x)$. The different choices of $\Psi$ correspond to different encoding schemes. 
We now construct a quantum operation that corresponds to a single ridgelet unit. For given parameters $(a,b,c)$, we define the unitary operator
\[ U_{a,b,c}(x) := R_{y} \!\left( 2c\,\eta_{a,b,u}(x) \right) = R_{y} \!\left( 2c\,a^{-1/2}\, \eta\!\left(\frac{x \cdot u-b}{a}\right) \right). \]
Here, the Ridgelet transform $\eta_{a,b}(x)$ determines the rotation angle of the qubit. The factor of $2$ is a normalization that simplifies later expressions for expectation values. Further, to determine the effect of $U_{a,b,c}(x)$, we apply it to the quantum state 
$|0\rangle$:
\begin{align*}
|\eta_{\text{out}}(x)\rangle = U_{a,b,c}(x)\,|0\rangle = \cos\!\Big( c\,a^{-1/2}\,\eta\!\left( \frac{x \cdot u-b}{a} \right) \Big)\,|0\rangle \\+  \sin\!\Big( c\,a^{-1/2}\,\eta\!\left( \frac{x \cdot u-b}{a} \right) \Big)\,|1\rangle. 
\end{align*}
Thus the ridgelet value appears inside trigonometric functions, which introduces a smooth, bounded nonlinearity at the quantum level. From this state, the probability of measuring $|1\rangle$ in the computational basis is written as    
\[ p_{1}(x) = \big|\langle 1|\eta_{\text{out}}(x)\rangle\big|^{2} = \sin^{2}\!\Big(c\,a^{-1/2}\,\eta\!\left( \frac{x \cdot u-b}{a} \right)\Big). \] Further, we measure the Pauli $Z$ gate, 
\(\ Z = \begin{pmatrix} 1 & 0\\[1mm] 0 & -1 \end{pmatrix}, \) its expectation value in the state $|\eta_{\text{out}}(x)\rangle$ is 
\(\ \langle Z\rangle_{x} = \langle \eta_{\text{out}}(x) | Z | \eta_{\text{out}}(x) \rangle = \cos\!\Big( 2c\,a^{-1/2}\, \eta\!\left(\frac{x \cdot u-b}{a}\right) \Big). \) Consequently, the classical function $x \mapsto \eta_{a,b,u}(x)$ is mapped to a bounded nonlinear function $x \mapsto \cos(2c\,a^{-1/2}\eta_{a,b,u}(x))$. This realizes a quantum analogue of a single nonlinear neuron. 
The computation of a classical ridgelet neural network with $J$ units is expressed as  
\[ g_{_J}(x) = \sum_{i=1}^{J} c_{i}\, a_{i}^{-1/2}\, \eta\!\left(\frac{x \cdot u_{i}-b_{i}}{a_{i}}\right). \] Here, each term in the sum represents the contribution of one ridgelet neuron with parameters $(a_{i}, b_{i}, c_{i})$. To construct the corresponding quantum model, we now apply a sequence of $y$--axis rotations, each associated with one ridgelet unit: 
\[ U_{J}(x) = \prod_{i=1}^{J} R_{y} \!\left( 2 c_{i}\, a_{i}^{-1/2}\, \eta\!\left(\frac{x  \cdot u_{i}-b_{i}}{a_{i}}\right) \right). \] 
Since all these gates are rotations about the same axis, they commute, and their angles are determined by the sum of each of their angles. More precisely, we have
\(\ R_{y}(\alpha_1)\,R_{y}(\alpha_2) = R_{y}(\alpha_1 + \alpha_2), \)
and therefore \[ U_{J}(x) = R_{y}\!\left( 2 \sum_{k=1}^{J} c_{i}\,a_{i}^{-1/2}\, \eta\!\left(\frac{x  \cdot u_{i}-b_{i}}{a_{i}}\right) \right) = R_{y}\!\left(2 g_{_J}(x)\right). \] Thus, the entire sequence of $J$ quantum ridgelet gates collapses into a single effective rotation whose angle is proportional to the classical ridgelet network output $g_{_J}(x)$. Applying $U_{J}(x)$ to $|0\rangle$ gives the final quantum state
\begin{align*}
   |\eta_{\text{out}}(x)\rangle = U_{J}(x)\,|0\rangle = R_{y}\!\left(2 g_{_J}(x)\right)\!|0\rangle \qquad\\  = \cos\!\left(g_{_J}(x)\right)|0\rangle + \sin\!\left(g_{_J}(x)\right)|1\rangle.   
\end{align*}

Hence, the amplitudes of the output state encode the value of the classical ridgelet network $g_{_J}(x)$ through trigonometric functions. We further consider a measurement of the Pauli $Z$ gate; the corresponding expectation value is
\(\ \langle Z\rangle^{J}_{x} = \langle \eta_{\text{out}}(x) | Z | \eta_{\text{out}}(x) \rangle = \cos\!\left(2 g_{_J}(x)\right). \) Therefore, the map \[ x \longmapsto \langle Z\rangle^{J}_{x} \] is a nonlinear transformation of the classical ridgelet network output. This hybrid structure features a single-qubit coupled with a ridgelet neural network for processing time series data.
The performance comparison presented in Table \ref{tab:Q1} demonstrates the effectiveness of our proposed QRNN model against both quantum and classical deep learning models, including QLSTM, QGRU, QTransformer, CRNN, LSTM, GRU, and Transformer. From the results, it is evident that QRNN achieves the best overall performance, with the lowest training RMSE and MAE, as well as the lowest testing RMSE and MAE. This does not indicate only high predictive accuracy but also strong generalization capability. In contrast, other quantum and classical deep learning models exhibit significantly higher error values across both training and testing phases, reflecting poorer learning and generalization performance. Furthermore, we tested our proposed model with a single-qubit system with various stocks from various sectors, and we can see that our model provides better outcomes in terms of accuracy and stability, as shown in Table \ref{tab:Q11}.

\begin{table*}[ht]
\centering
\renewcommand{\arraystretch}{5}
\rowcolors{2}{blue!10}{blue!10} % Apply light blue to data rows
    \renewcommand{\arraystretch}{1.2} % Adjust row height

\begin{tabular}{|l|c|c|c|c|}
\rowcolor{green!30} % Header row in green
\toprule
\textbf{Model} & \textbf{Training RMSE} & \textbf{Training MAE} & \textbf{Testing RMSE} & \textbf{Testing MAE} \\
\midrule
QRNN          &  0.018573  &   0.015641 & {0.040137}  & { 0.029451}  \\
QLSTM \cite{li2024quantum}        & 0.100942  & 0.087337   &  0.405184    &   0.385651  \\
QGRU \cite{li2024quantum}         &  0.081458  &  0.070452  & 0.334041  &  0.316869 \\
QTransformer \cite{zhang2025quantum}  & 0.076142  & 0.065620   & 0.370580 &  0.347840  \\
CRNN          &  0.010275  &  0.007784   & 0.086217  &  0.070392 \\
LSTM          & 0.089625  & 0.073255  & 0.214902  &   0.197111 \\
GRU           & 0.018507  & 0.015374  &   0.185229 &  0.163096 \\
Transformer   &   0.007895 &  0.005053 &  0.229551  &   0.201858 \\
\bottomrule
\end{tabular}
\caption{Prediction errors between QRNN, QLSTM, QGRU, QTransformer   CRNN, LSTM, GRU, and Transformer with single-qubit system.}
\label{tab:Q1}
\end{table*}

\begin{table*}[ht]
\centering
\renewcommand{\arraystretch}{5}
\rowcolors{2}{blue!10}{blue!10} % Apply light blue to data rows
    \renewcommand{\arraystretch}{1.2} % Adjust row height

% \begin{adjustbox}{max width=\textwidth}
\begin{tabular}{|p{3cm}|p{1.5cm}|p{1.5cm}|p{1.8cm}|p{1.6cm}|}
% \begin{tabular}{lrrrrr}
\rowcolor{green!30} % Header row in green
\toprule
{Stock} & {Phase} & { Metric} & {CRNN} & {QRNN}   \\
\midrule
APPLE      & Test & RMSE &  0.105341 &  { 0.043391}    \\
           &       & MAE  &0.081971  &    { 0.031096}\\
\midrule           
 SONY      
           & Test  & RMSE &0.110905 & {0.053212}\\
  
        &       & MAE &  0.100325& {0.039742  }\\
\midrule
TESLA     
           
           & Test  & RMSE& 0.083813& 0.066843{ }\\
           &       & MAE & 0.041538  & {0.034625 }\\
\midrule
NVIDIA   & Test  & RMSE&  0.206544  & {0.070897}  \\
           &       & MAE & 0.177895  & {0.051273 }  \\
\midrule
INTEL      
           & Test  & RMSE& 0.122567 & {0.059632}  \\
           &       & MAE &  0.102202 & {0.042452}  \\
\midrule
GENERAL MOTOR   & Test  & RMSE&  0.050316& { 0.032153}  \\
           &       & MAE &0.038451 &  {0.023523} \\
\midrule
FORD    & Test  & RMSE&0.039106 & { 0.035506}  \\
           &       & MAE  &0.030748 & {  0.027004}  \\
\midrule

AAL     
           & Test  & RMSE&  0.052535& {0.048991}  \\
           &       & MAE &0.038934 & {  0.030055  }  \\
\midrule
 AMD    
           & Test  & RMSE&0.102331  & {0.078805 }\\
           &       & MAE &0.069866  & { 0.048614 }  \\
\midrule
GOOGLE     
           & Test  & RMSE& 0.110905  & {0.042135 }\\
           &       & MAE & 0.092325& {0.031167}  \\
\midrule
AMAZON    
           & Test  & RMSE& 0.123007  & {0.045212 }\\
           &       & MAE & 0.104958 & {0.032756 }  \\
\bottomrule
\end{tabular}
\caption{Performance comparison of the proposed method compared to the baseline model with a single-qubit system.}
\label{tab:Q11}
\end{table*}

\subsection{QRNN for multi- qubit system}

In this subsection, we present the generalization of the proposed QRNN model for an $n$-qubit system. 
Here $\mathbb{R}$ represents the set of real numbers and $\mathbb{C}$ denotes the set of complex numbers. The state space of an $n$-qubit quantum system represents the Hilbert space
\(\
\mathcal{H}_n := (\mathbb{C}^2)^{\otimes n},
\)
which is a $2^n$-dimensional complex vector space.
The computational basis of $\mathcal{H}_n$ is denoted as
\(\
\big\{\,|i_1 i_2 \cdots i_n\rangle \;:\; i_j \in \{0,1\}, \; j=1,\dots,n \,\big\},
\)
where each basis vector is a classical binary configuration of the qubits.
Now, we define the Pauli operator $Y$, which is denoted by $\sigma_y$ , and defined as
\[
\sigma_y =
\begin{pmatrix}
0 & -i \\
i & 0
\end{pmatrix}.
\]
The single-qubit rotation operator on the $y$-axis is represented by $R_y(\theta)$ and is defined as
\[
R_y(\theta) := e^{-i \theta \sigma_y / 2}
=
\begin{pmatrix}
\cos(\theta/2) & -\sin(\theta/2) \\
\sin(\theta/2) & \cos(\theta/2)
\end{pmatrix}.
\]
Here, $\theta \in \mathbb{R}$ denotes the rotation angle.
For $j = 1,\dots,n$, we denote by $R_y^{(j)}(\theta)$ the operator that applies the rotation $R_y(\theta)$ to the $j$-th qubit and acts as identity on all other qubits.
Let $x \in \mathbb{R}^d$ be a classical input vector.
Let $\Psi_j : \mathbb{R}^d \to \mathbb{R}$, for $j=1,\dots,n$, be real-valued encoding functions. We now consider the initial quantum state
\[
|0\rangle^{\otimes n}
=
|0\rangle \otimes \cdots \otimes |0\rangle \in \mathcal{H}_n.
\]
The encoded quantum state is defined by applying single-qubit rotations to each qubit and is written as
\[
|\eta_{\mathrm{in}}(x)\rangle
:=
\left(
\prod_{j=1}^{n}
R_y^{(j)}(\Psi_j(x))
\right)
|0\rangle^{\otimes n}.
\]
Using the explicit form of the rotation operator, we get the following
\[
|\eta_{\mathrm{in}}(x)\rangle
=
\bigotimes_{j=1}^{n}
\left[
\cos\!\left(\frac{\Psi_j(x)}{2}\right)|0\rangle
+
\sin\!\left(\frac{\Psi_j(x)}{2}\right)|1\rangle
\right].
\]
Let $\eta : \mathbb{R} \to \mathbb{R}$ be a ridgelet function. For parameters $a > 0$ and $b \in \mathbb{R}$, we define the ridgelet family
\[
\eta_{a,b,u}(x)
:=
a^{-1/2}\,
\eta\!\left(\frac{x \cdot u-b}{a}\right).
\]
For each qubit $j$, we define a finite ridgelet neural network expansion
\[
g_j(x)
:=
\sum_{k=1}^{J}
c_{j,k}\, a_{j,k}^{-1/2}\,
\eta\!\left(\frac{x \cdot u_{j,k}-b_{j,k}}{a_{j,k}}\right),
\]
where
\(\
c_{j,k} \in \mathbb{R}, \quad a_{j,k} > 0, \quad b_{j,k} \in \mathbb{R}, \quad u_{j,k} \in \mathbb{R}^d.
\)
For each qubit $j$ and each term $k$, define the angle
\[
\theta_{j,k}(x)
:=
2 c_{j,k}\, a_{j,k}^{-1/2}\,
\eta\!\left(\frac{x \cdot u_{j,k}-b_{j,k}}{a_{j,k}}\right).
\]
We consider the ridgelet expansion as generating rotation angles of quantum gates. This establishes a connection between the classical ridgelet representation and parameterized quantum processes.
Therefore, the quantum ridgelet operator associated with the ridgelet expansion is defined as
\[
U_J(x)
:=
\prod_{j=1}^{n}
\prod_{k=1}^{J}
R_y^{(j)}\big(\theta_{j,k}(x)\big).
\]
We employ the identity that rotations about the same axis commute and satisfy the additive property:
\(\
R_y(\theta_1) R_y(\theta_2)
=
R_y(\theta_1 + \theta_2).
\)
By induction, we obtain the following
\[
\prod_{k=1}^{J}
R_y(\theta_k)
=
R_y\!\left(\sum_{k=1}^{J} \theta_k\right).
\]
Applying this to each qubit, we get the following
\[
\prod_{k=1}^{J}
R_y^{(j)}(\theta_{j,k}(x))
=
R_y^{(j)}\!\left(
\sum_{k=1}^{J} \theta_{j,k}(x)
\right).
\]
By applying the definition of $g_j(x)$, this becomes
\[
\sum_{k=1}^{J} \theta_{j,k}(x)
=
2 g_j(x).
\]
Hence,
\(\
U_J(x)
=
\prod_{j=1}^{n}
R_y^{(j)}\!\left(2 g_j(x)\right).
\)
We now define the output quantum state as
\(\
|\psi_{\mathrm{out}}(x)\rangle
:=
U_J(x)\,|0\rangle^{\otimes n}.
\)
For each qubit $j$, the operator $R_y^{(j)}(2 g_j(x))$ acts on the $j$-th component of the tensor product state. Using the single-qubit identity
\(\
R_y(2\theta)\,|0\rangle
=
\cos(\theta)|0\rangle + \sin(\theta)|1\rangle.
\)
We obtain that each qubit transforms as
\(\
|0\rangle \;\mapsto\;
\cos\!\left(g_j(x)\right)|0\rangle
+
\sin\!\left(g_j(x)\right)|1\rangle.
\)
Therefore,
\[
|\eta_J(x)\rangle
=
\bigotimes_{j=1}^{n}
\left[
\cos\!\left(g_j(x)\right)|0\rangle
+
\sin\!\left(g_j(x)\right)|1\rangle
\right].
\]
Let $Z = \begin{pmatrix}1 & 0 \\ 0 & -1\end{pmatrix}$ represent the Pauli-$Z$ operator, and let $Z_j := Z^{(j)}$ denote its action on the $j$-th qubit.
We compute
\(\
\langle Z_j \rangle_x
:=
\langle \eta_{\mathrm{out}}(x) | Z_j | \eta_{\mathrm{out}}(x) \rangle.
\)
Using the product structure of the state and the single-qubit result, we obtain
\(\
\langle Z_j \rangle_x
=
\cos\!\left(2 g_j(x)\right).
\)
Further, we define the joint observable
\(\
Z^{\otimes n} := Z \otimes \cdots \otimes Z.
\) Then
\[
\langle Z^{\otimes n} \rangle_x
=
\prod_{j=1}^{n}
\cos\!\left(2 g_j(x)\right).
\]
If
\(\
g_j(x) = g_J(x), \quad \forall j,
\)
then we have
\(\
U_J(x)
=
\prod_{j=1}^{n}
R_y^{(j)}\!\left(2 g_J(x)\right),
\)
and
\(\
\langle Z^{\otimes n} \rangle_x
=
\cos^n\!\left(2 g_J(x)\right).
\)

 The computation of the quantum-classical ridgelet neural network model is discussed in Algorithm \ref{algorithm}.

\begin{algorithm*}[!t]
\caption{Computation of Quantum Ridgelet Neural
Network Model}
\KwIn{Time series data $X$, input dimension $e$, number of ridgelet features $m_e$, number of qubits $m_q$, parameter $\theta$}
\KwOut{Predicted outputs $\hat{Y}_{\text{pred}}$}

\SetKwFunction{FHybridModel}{ QRNN Model}
\SetKwProg{Fn}{Procedure}{:}{}
\Fn{\FHybridModel{X}}{
    1. Initialize ridgelet feature  transformer with time series data $X$ and input dimension $e$ \;
    2. Initialize linear map from ridgelet feature transformer to qubit space of size $m_q$\;
    3. Construct quantum circuit module: {QuantumCircuitModule}($m_q$, shots, $\theta$)\;
    4. Initialize final prediction block (classical fully connected layers)\;
    5. Initialize empty list for quantum outputs: $Q\_outs \gets [~]$\;
    6. \ForEach{sample $x_i$ in $X$}{
        a. Compute ridgelet features $h_i \gets {RidgeletFeature}(x_i)$\;
        b. Map features to qubit space: $w_i \gets {Linear}(h_i)$\;
        c. Compute quantum outputs: $Q_i \gets {Quantum Circuit Module.forward\_pass}(w_i, \theta)$\;
        d. Append $Q_i$ to $Q\_outs$\;
    }
    7. Stack quantum outputs: $Q\_out \gets {stack}(Q\_outs)$\;
    8. Generate final prediction: $\hat{Y}_{\text{pred}} \gets {block}(Q\_out)$\;
    9. \textbf{Return} $\hat{Y}_{\text{pred}}$\;
}
\label{algorithm}
\end{algorithm*}

\section{QAOA-Based Portfolio Optimization}\label{section:5}

Here, based on the unified portfolio optimization problem in the previous section \ref{section:3}, our proposed QRNN model generated the expected return vector, which is employed to perform quantum-enhanced portfolio optimization via the QAOA.  The goal of this step is to use a variational quantum optimization framework to solve the constrained binary portfolio selection problem.

To enable quantum optimization via QAOA, the cardinality-constrained portfolio optimization problem is transformed into a QUBO formulation by incorporating the cardinality constraint into the objective function through a quadratic penalty term. The following is the expression for the resulting QUBO objective:

\begin{equation*} Q(x)= -\hat{\mu}^{\top}x+x^{\top}\Sigma x + P\left(\sum_{i=1}^{N}x_i-K\right)^2, \end{equation*} where $P>0$ is a penalty coefficient that enforces the cardinality constraint.

Furthermore, the QUBO formulation is subsequently mapped to an Ising-type cost Hamiltonian $H_C$, which serves as the optimization Hamiltonian for QAOA. The QAOA is a variational quantum optimization algorithm designed to solve combinatorial optimization problems through parameterized quantum circuits. QAOA prepares the variational quantum state, which is expressed as follows: \begin{equation*} |\psi(\boldsymbol{\gamma},\boldsymbol{\beta})\rangle = U_M(\boldsymbol{\beta}) U_C(\boldsymbol{\gamma}) |+\rangle^{\otimes N}, \end{equation*} where \begin{equation*} U_C(\boldsymbol{\gamma}) = \exp(-i\gamma H_C) \end{equation*} is the cost unitary associated with the problem Hamiltonian, and \begin{equation*} U_M(\boldsymbol{\beta}) = \exp(-i\beta H_M) \end{equation*} is the mixing unitary associated with the mixer Hamiltonian $H_M$. The optimal variational parameters are obtained by minimizing the expected cost Hamiltonian value: \begin{equation*} (\boldsymbol{\gamma}^{*},\boldsymbol{\beta}^{*}) = \arg\min_{\boldsymbol{\gamma},\boldsymbol{\beta}} \left\langle \psi(\boldsymbol{\gamma},\boldsymbol{\beta}) \middle| H_C \middle| \psi(\boldsymbol{\gamma},\boldsymbol{\beta}) \right\rangle. \end{equation*} After variational optimization, repeated measurement of the optimized quantum state yields candidate binary portfolio vectors, and the solution with the minimum QUBO value is selected as the optimal portfolio allocation. The resulting optimal binary vector $x^{*}$ identifies the final selected portfolio, where $x_i^{*}=1$ indicates that asset $i$ is included in the optimized portfolio. The computation mechanism of QRNN-QAOA portfolio optimization is discussed in Algorithm \ref{alg:qrnn_qaoa}.

\begin{algorithm*}[t] \caption{Integrated QRNN-QAOA Portfolio Optimization Framework} \SetKwFunction{FHybridModel}{QRNN-QAOA Portfolio Optimization Framework} \SetKwProg{Fn}{Procedure}{:}{} \Fn{\FHybridModel{$X, K$}}{ \textbf{Input:} Historical stock market data $X$, desired portfolio cardinality $K$\; \textbf{Output:} Predicted return vector $\hat{\mu}$, optimized portfolio selection vector $x^{*}$\; 1. Compute stock return series from historical market data $X$\; 2. Apply ridgelet transform denoising to the return series\; 3. Construct sliding-window input sequences from denoised returns\; 4. Initialize ridgelet feature transformer with input dimension $e$ and $m_e$ ridgelet directions\; 5. Initialize linear mapping from ridgelet feature space to qubit space of size $m_q$\; 6. Construct quantum circuit module: {QuantumCircuitModule}($m_q$, shots, $\theta$)\; 7. Initialize final prediction block (classical fully connected layers)\; 8. Initialize empty list for quantum outputs: $q\_outs \gets [~]$\; 9. \ForEach{sample $x_i$ in sliding-window sequences}{ a. Compute ridgelet features: $h_i \gets {RidgeletFeature}(x_i)$\; b. Map ridgelet features to qubit space: $w_i \gets {Linear}(h_i)$\; c. Compute quantum outputs: $Q_i \gets {QuantumCircuitModule.forward\_pass}(w_i,\theta)$\; d. Append $Q_i$ to $q\_outs$\; } 10. Stack quantum outputs: $q\_out \gets {stack}(q\_outs)$\; 11. Generate predicted expected return vector: $\hat{\mu} \gets {PredictionBlock}(q\_out)$\; 12. Estimate covariance matrix of historical asset returns: $\Sigma \gets {Covariance}(X)$\; 13. Formulate portfolio optimization objective: $\min -\hat{\mu}^{\top}x + x^{\top}\Sigma x$\; 14. Impose portfolio cardinality constraint: $\sum_{i=1}^{N}x_i = K$\; 15. Convert the constrained optimization problem into QUBO form\; 16. Initialize QAOA optimizer with QUBO Hamiltonian\; 17. Solve QUBO using QAOA to obtain optimal binary portfolio vector: $x^{*}$\; 18. Select portfolio assets corresponding to: $x_i^{*}=1$\; 19. \textbf{Return} \text{optimized portfolio:} $\hat{\mu},  x^{*}$\; } \label{alg:qrnn_qaoa}
\end{algorithm*}

 \section{Quantum Error Correction Analysis}\label{section:6}
In this section, we further analyze the robustness of our proposed QRNN model under different noise and error mitigation conditions. It is crucial to determine how our proposed QRNN model operates in the presence of quantum noise and how error correction approaches enhance the productivity of our model. With this objective, we examine four distinct variants of our proposed QRNN model. First, we analyze the noise-free quantum model, which introduces quantum without any noise or correction technique. Second, in order to figure out how noise affects our proposed QRNN model stability, we further investigate its effects under different scenarios. Third, we employ resilience and corrective techniques to reduce errors by incorporating QEC into our noisy quantum model.
 Lastly, we examine a partial QEC technique that reduces complexity without compromising robustness by applying only certain parts of the error-correction operations.

\subsection{Parametrized Variational Circuit} 
\begin{figure*}[!ht]
    \centering
    % --------- Second figure ---------
    \includegraphics[width=14cm]{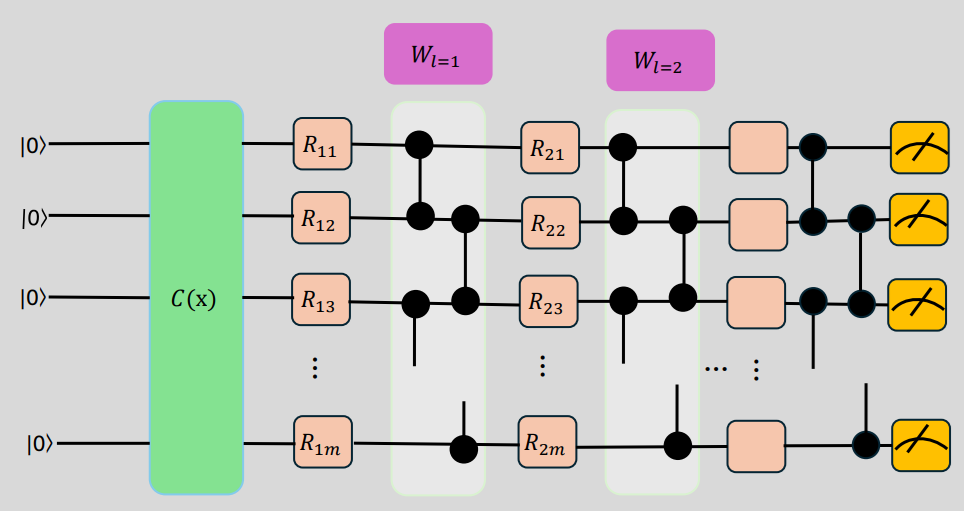}
    \caption{The detailed circuit diagram of the VQC depicted in this paper. The data is encoded into the circuit through angle encoding. The parametrized unitary consists of multiple layers of single-qubit rotations, indicated as  $R_{lm}$, followed by entangling CNOT gates $W_l$, and measurements are performed using Pauli-Z observables.
}
     \label{fig:3}
 \end{figure*}

A parametrized variational circuit is a critical component of our quantum ridgelet neural networks. First, the characteristic of the input data $x_i$ is encoded into a quantum state,  which is expressed as follows \begin{equation*}
|\Psi(x_i)\rangle = C(x_i)\, |0\rangle^{\otimes n}. \end{equation*}
Second, the state evolves through a parametrized unitary transformation $U(\theta)$ with parameters $\theta \in [-\pi, \pi]^D$,  which is decomposed into a sequence of $D$ unitaries with layer $l$ on the qubit $n$ which is written as 
\begin{equation*} U(\theta) = \prod_{l=1}^{D} \left( W_l \prod_{m=1}^{n} e^{-i \theta_{lm} H_{lm}} \right), \end{equation*}
where the training parameters are $\theta_{lm}$ and the single-qubit rotation generators are $H_{lm} = \hat{n}_{lm} \cdot \sigma$, where $\sigma = (X, Y, Z)^T$. Entangling layers without trainable parameters are represented by the operators $W_l$. Each rotation gate $R_{lm}:= e^{-i \theta_{lm} H_{lm}}$ with three degrees of freedom can be represented in terms of Euler angles \(\alpha, \beta, \gamma\): \begin{equation*} R_{lm} = R_Z(\alpha_{lm}) R_Y(\beta_{lm}) R_Z(\gamma_{lm}). \end{equation*} Here, we use a hardware-efficient ansatz consisting of single-qubit rotations and entangling gates to implement the parametrized unitary. In the final stage of the circuit, measurements are performed using Pauli-$Z$ observables: \begin{equation*} y_i = \langle \Psi(\theta) | Z_i | \Psi(\theta) \rangle. \end{equation*} The circuit design is illustrated in Fig.~\ref{fig:3}.

\subsection{Quantum Noise Configuration} 
Here, to account for noise effects in the PQC, we consider a combination of standard single-qubit noise channels, including bit-flip, phase-flip, and depolarizing noise. For a quantum state $\rho$, the bit-flip channel is expressed as \begin{equation*} \mathcal{E}_X(\rho) = (1 - p)\rho + p X \rho X, \end{equation*} the phase-flip channel is written as \begin{equation*} \mathcal{E}_Z(\rho) = (1 - p)\rho + p Z \rho Z, \end{equation*} and the depolarizing channel is represented by \begin{equation*} \mathcal{E}_{\text{depol}}(\rho) = (1 - p)\rho + \frac{p}{3}\left( X \rho X + Y \rho Y + Z \rho Z \right), \end{equation*} where $p \in [0,1]$ represents the noise strength, and $X, Y, Z$ denote the Pauli operators. In our framework, these noise channels are applied sequentially after each quantum operation, resulting in a combined noise process \begin{equation*} \mathcal{E} = \mathcal{E}_{\text{depol}} \circ \mathcal{E}_Z \circ \mathcal{E}_X. \end{equation*} Here, $U_l(\theta_l)$ represent the unitary operation at layer $l$. The overall noisy evolution of an input state $\rho_{\text{in}}$ through a circuit of depth $L$ is expressed as in the following\begin{equation*} \rho_{\text{out}} = \mathcal{E}_L \Big( U_L(\theta_L) \cdots \mathcal{E}_1 \big( U_1(\theta_1)\rho_{\text{in}} U_1^\dagger(\theta_1) \big) \cdots U_L^\dagger(\theta_L) \Big), \end{equation*} where $\mathcal{E}_l$ denotes the combined noise channel applied at layer $l$. For a circuit with $N$ noisy operations, the cumulative effect of noise is approximated by the following \begin{equation*} \epsilon_{\text{total}} \approx 1 - (1 - p)^N. \end{equation*} This depicts how noise increases with circuit depth. This framework demonstrates a systematic analysis of noise effects on the performance of the PQC within a simulation-based environment. The flow diagram is displayed in Fig. \ref{fig:Noise}.

\subsection{Quantum Error Correction Configuration} Here, quantum error correction is incorporated to mitigate the impact of noise. We employ a three-qubit repetition code to encode each logical qubit into three physical qubits. The logical basis states are given as in the following: \begin{equation*} |0_L\rangle = |000\rangle, \quad |1_L\rangle = |111\rangle. \end{equation*} For an arbitrary single-qubit state \
\(\ |\psi\rangle = \alpha |0\rangle + \beta |1\rangle\),  the corresponding encoded logical state is given by \[\ |\psi_L\rangle = \alpha |000\rangle + \beta |111\rangle, \] where $\alpha, \beta \in \mathbb{C}$ satisfies $|\alpha|^2 + |\beta|^2 = 1$. The density matrix of the encoded logical state is represented as  \(\ \rho_L = |\psi_L\rangle \langle \psi_L|.\) The same noise model $\mathcal{E}$ is applied independently to each physical qubit, resulting in a noisy state  \(\ \rho_{\text{noisy}} = \mathcal{E}(\rho_L). \)  After noise implementation, an error correction operation $\mathcal{R}$ is performed to recover the logical state: \(\  \rho_{\text{corr}} = \mathcal{R}(\rho_{\text{noisy}}). \)  In practice, $\mathcal{R}$ is applied through pairwise quantum operations, which approximates a voting mechanism to correct bit-flip errors. The flow diagram is depicted in Fig. \ref{fig:Noise}.

 \subsection{Partial Quantum Error Correction Configuration}
\begin{figure*}[!ht]
    \centering
    % --------- Second figure ---------
    \includegraphics[width=14cm]{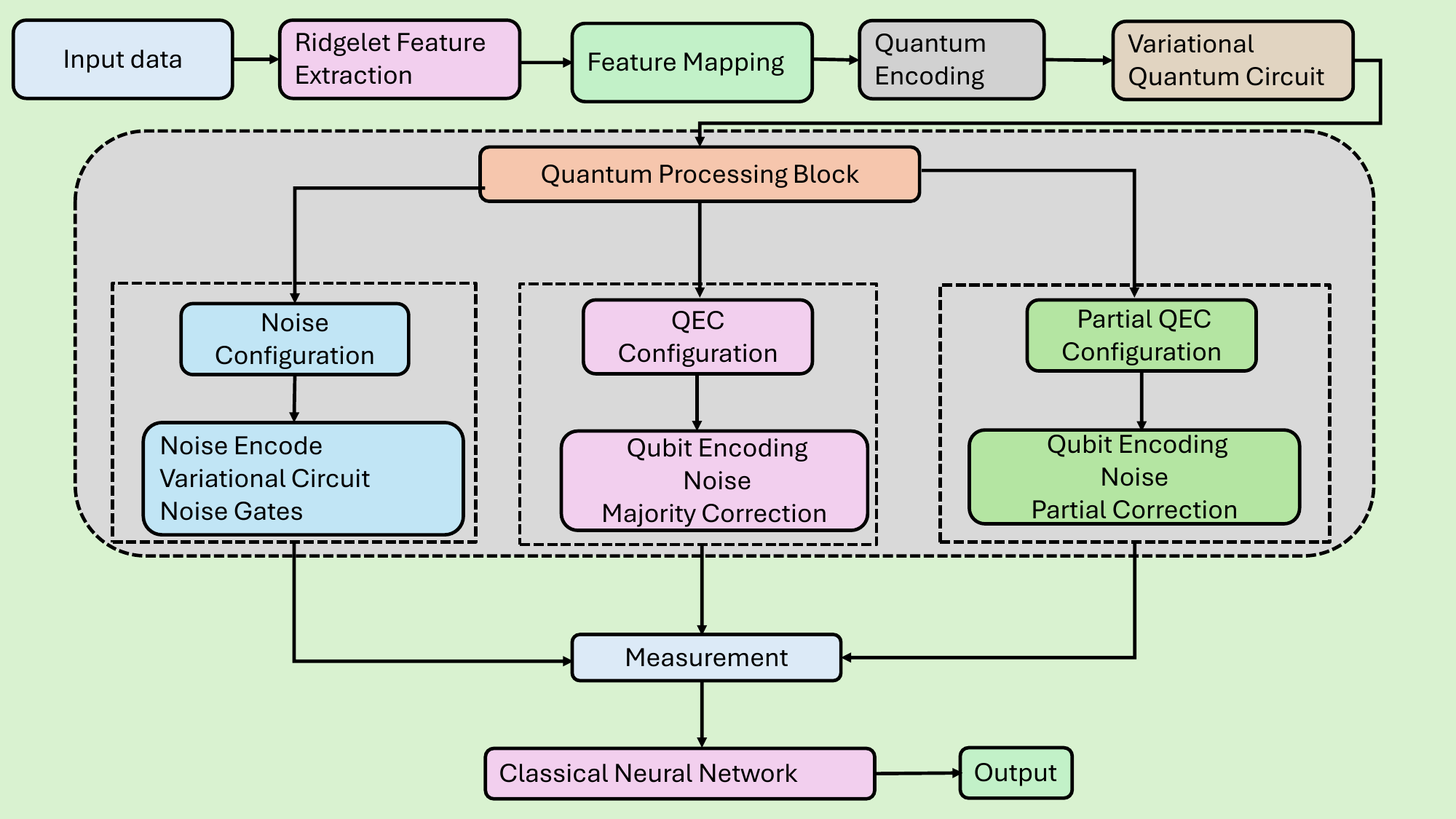}
    \caption{ Systematic architecture of quantum error correction circuit.
}
     \label{fig:Noise}
 \end{figure*}

In order to reduce computational complexity while preserving noise mitigation, a partial QEC method is applied in this configuration.  Each logical qubit is encoded using the same three-qubit repetition pattern similar to the entire QEC configuration: \(\  |\psi_L\rangle = \alpha |000\rangle + \beta |111\rangle. \) The corresponding density matrix is given by \(\  \rho_L = |\psi_L\rangle \langle \psi_L|.\) The same noise channel $\mathcal{E}$ is applied to all physical qubits: \(\ \rho_{\text{noisy}} = \mathcal{E}(\rho_L). \) However, instead of applying full recovery, a reduced correction operation $\mathcal{R}_{\text{partial}}$ is used: \(\  \rho_{\text{corr}} = \mathcal{R}_{\text{partial}}(\rho_{\text{noisy}}), \) where $\mathcal{R}_{\text{partial}}$ performs a limited correction using a subset of the available redundancy. This approach provides partial noise mitigation while maintaining lower circuit complexity, enabling analysis of the trade-off between correction performance and computational efficiency. The flow diagram is illustrated in Fig. \ref{fig:Noise}.

The performance of our QRNN model under various variations for single, double, and multi-qubit scenarios is shown in Tables \ref{tab:single qubit}, \ref{tab:double qubit}, and \ref{tab:multi qubit}.  The comparison includes our proposed QRNN model, the noisy version, and the models with QEC and partial QEC are given in Tables \ref{tab:single qubit}, \ref{tab:double qubit}, and \ref{tab:multi qubit}. 

Our proposed QRNN model obtains the lowest error for the single-qubit scenario as indicated in Table \ref{tab:single qubit}, suggesting that our model works best in a perfect, noise-free environment. The inaccuracy significantly rises when noise is included, indicating the considerable influence of quantum noise on our model performance. Appropriate noise mitigation is demonstrated by the partial QEC setup, which lowers the error. However, when compared to partial QEC, a complete QEC approach produces a marginally larger inaccuracy, most likely as a result of increased quantum circuit complexity, as shown in Table \ref{tab:single qubit}.

\begin{table*}[ht]
\centering
\renewcommand{\arraystretch}{5}
\rowcolors{2}{blue!10}{blue!10} % Apply light blue to data rows
    \renewcommand{\arraystretch}{1.2} % Adjust row height

\begin{tabular}{|l|c|c|c|c|}
\rowcolor{green!30} % Header row in green
\toprule
Metric & QRNN(without noise) & QRNN(with Noise) & QRNN(QEC) & QRNN( Partial QEC) \\
\midrule
RMSE &  0.071367 &   0.298718  &  0.235277 &   0.217090     \\
\midrule
MAE  & 0.058005  &  0.282329 &   0.233431  &  0.204468 \\

\bottomrule
\end{tabular}
\caption{Performance comparison of the proposed method compared to different quantum variants of QRNN with a single-qubit.}
\label{tab:single qubit}
\end{table*}

\subsection{Multi Qubit Analysis}

Here, we increase the number of qubits for further investigation in order to see the behavior of our proposed model under these different configurations. In the two-qubit scenario, as indicated in Table \ref{tab:double qubit}, we found that the noisy model exhibits a significant decline in performance, whereas our proposed QRNN model performs well but not better than a single-qubit. While full QEC produces a moderate improvement but is still less effective than partial QEC.
As seen in Table \ref{tab:multi qubit}, the total error values for the multi-qubit situation rise as the system complexity increases. The noisy model performs the poorest, whereas the baseline QRNN succeeds. Furthermore, partial QEC enhances the outcomes, whereas full QEC offers some degree of adjustment. The difficulties of scaling our quantum models under noise are highlighted by the fact that, while error correction aids, the benefit is not as noticeable as it would be in lower qubit environments.

\begin{table*}[ht]
\centering
\renewcommand{\arraystretch}{5}
\rowcolors{2}{blue!10}{blue!10} % Apply light blue to data rows
    \renewcommand{\arraystretch}{1.2} % Adjust row height

\begin{tabular}{|l|c|c|c|c|}
\rowcolor{green!30} % Header row in green
\toprule
Metric & QRNN(without noise) & QRNN(with Noise) & QRNN(QEC) & QRNN( Partial QEC) \\
\midrule

RMSE & 0.052879 & 0.177092  & 0.120839  & 0.116233     \\
\midrule
MAE  &  0.038205   &  0.172549& 0.114123  &  0.109243   \\

\bottomrule
\end{tabular}
\caption{Performance comparison of the proposed method compared to different quantum variants of QRNN with a double qubit.}
\label{tab:double qubit}
\end{table*}

\begin{table*}[ht]
\centering
\renewcommand{\arraystretch}{5}
\rowcolors{2}{blue!10}{blue!10} % Apply light blue to data rows
    \renewcommand{\arraystretch}{1.2} % Adjust row height

\begin{tabular}{|l|c|c|c|c|}
\rowcolor{green!30} % Header row in green
\toprule
Metric & QRNN(without noise) & QRNN(with Noise) & QRNN(QEC) & QRNN( Partial QEC) \\
\midrule
RMSE & 0.032263   & 0.113731  & 0.092420    &   0.074563   \\
\midrule
MAE  &  0.023578    &  0.106940 &   0.085496 &    0.066922  \\

\bottomrule
\end{tabular}
\caption{Performance comparison of the proposed method compared to different quantum variants of QRNN with multiple qubits.}
\label{tab:multi qubit}
\end{table*}

\section{Experiment with Real World Data} \label{section:7}
\begin{figure*}[!ht]
    \centering
    % --------- Second figure ---------
    \includegraphics[width=14cm]{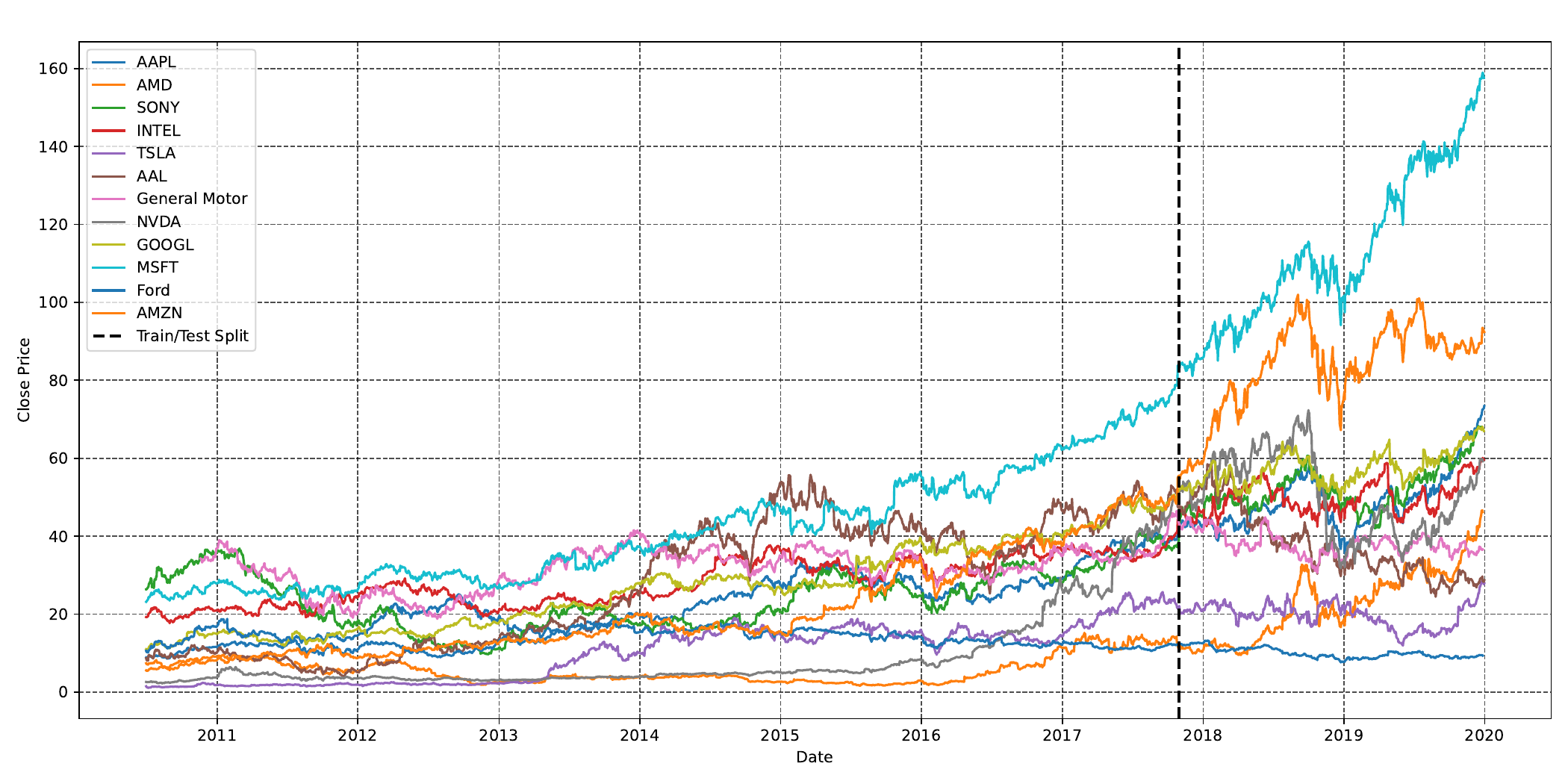}
    \caption{Stock Price Trends.
     \label{fig:4}
}
   
\end{figure*}

Here, we evaluated the past closing prices of 12 stocks from January 7, 2010, to December 31, 2019, for this study. We consider the following companies, each with a ticker symbol: AAPL, MSFT, AMZN, NVDA, GOOGL, TSLA, AAL, SONY, INTC, AMD, F, and GM. This information, which included past stock prices, was collected from Yahoo Finance.  This large data set provides a thorough picture of the stock's behavior over time by capturing a variety of market situations.
Effective forecasting requires models to learn both short-term volatility and long-term trends in the stock price sequence, which is made possible by the availability of long-term data. By making use of this time frame, we make sure that our models are trained on an extensive and diverse data set, which enhances their robustness and generalization accuracy for forecasting of stock prices. In order to stabilize training, specifically when integrating classical neural network components with quantum layers, normalizing all of these features to transfer their values to the range of 0 and 1. The data set was then divided into a 70\%  training set and a 30\%  testing set, as shown in Fig. \ref{fig:4}. Our experimental setup utilized 16 ridge directions for the ridgelet component, ensuring an accurate representation of the input feature space while effectively reducing overfitting. To enable a deep, yet highly computational quantum feature mapping, VQC was designed for the quantum module of the proposed model, utilizing 6 qubits and 6 repeating VQC layers of parameterized quantum gates. In order to guarantee stable convergence during optimization, the framework was trained employing mini-batches of size 32 across 50 epochs. By balancing learning stability and generalization, we avoided overfitting and encouraged smooth optimization during training by using a learning rate of \(5 \times 10^{-4}\)and a weight decay of \(1 \times 10^{-5}\). To handle the presence of variations in stock prices while maintaining uniform gradients for optimization, the Huber Loss function was employed with a parameter $\delta = 1.0$. The sequences were then transformed into PyTorch tensors and assigned to the High-Performance Computing (HPC) system, preparing the data for effective model training. To effectively train both classical and quantum-enhanced ridgelet models, all tests were conducted on an HPC cluster equipped with NVIDIA H100 NVL GPUs, which provided significant processing power and memory bandwidth. By employing these GPUs, we significantly accelerated quantum circuit simulators, matrix operations, and gradient computations, resulting in faster convergence and shorter training times. Additionally, using the NVIDIA H100 NVL GPUs architecture guaranteed better optimization dynamics and increased computational precision, which is especially advantageous for quantum variational layers that are sensitive to tiny gradient fluctuations.

\subsection{Evaluation Metrics}
To evaluate the performance of our model, we utilized two key evaluation metrics: Root Mean Squared Error (RMSE) and Mean Absolute Error (MAE). By taking the square root of the mean squared differences between actual and predicted values, the RMSE measures the average size of prediction mistakes. It gives more weight to models that infrequently generate significant errors, as it is especially sensitive to large variances. Instead of unfairly penalizing outliers, the MAE provides a straightforward understanding of the model's average prediction error by measuring the average absolute difference between predicted and actual values. The formulas are given as follows:
\begin{equation*}
\text{RMSE} = \sqrt{\frac{1}{N} \sum_{t=1}^{N} (Y_t - \hat{Y}_t)^2,}
\end{equation*}
\begin{equation*}
\text{MAE} = \frac{1}{N} \sum_{t=1}^{N} \left|Y_t - \hat{Y}_t \right|,
\end{equation*}

where $Y_t$ is the normalized true stock price value and $\hat{Y}_t$ is the normalized predicted value for the $t^{th}$ day.

\subsection{Accuracy and Loss}
The forecasting ability of our proposed QRNN model was evaluated through a comprehensive performance comparison with 
quantum and classical baseline models, including QLSTM, QGRU, QTransformer, CRNN, LSTM, GRU, and Transformer, on a wide range of stocks from various industries. RMSE and MAE, two common regression metrics that together evaluate prediction accuracy and robustness to variance, were used in this study. The prediction losses of QLSTM, QGRU, QTransformer, LSTM, GRU, Transformer, QRNN, and CRNN on training and test sets are displayed in Table \ref{tab:1}. The proposed QRNN model outperforms the baseline models in terms of prediction accuracy in extremely volatile market conditions, as indicated by lower RMSE and MAE values for both the training and test sets, as shown in Table \ref{tab:1}.  While classical models, such as LSTM, GRU, and Transformer, consistently performed lower in predicted accuracy, CRNN emerged as the second-best performer. This is reflective of their very limited ability to capture complicated nonlinear interactions in stock price prediction. However, CRNN emerged as the second-best performer. Notably, assets from several industries were included in the evaluation, guaranteeing that QRNN's performance is not limited to one sector or company but rather generalizes well across a variety of heterogeneous datasets. This wide range of applications highlights QRNN's resilience and dependability, especially under erratic and unpredictable market circumstances. The results demonstrate that QRNN is an innovative approach for stock market prediction tasks, as it not only achieves improved accuracy but also performs consistently across various market regimes, surpassing quantum and classical baseline models, as shown in Tables \ref{tab:1} and \ref{tab:2}.

\begin{table*}[ht]
\centering
\renewcommand{\arraystretch}{5}
\rowcolors{2}{blue!10}{blue!10} % Apply light blue to data rows
    \renewcommand{\arraystretch}{1.2} % Adjust row height

\begin{tabular}{|l|c|c|c|c|}
\rowcolor{green!30} % Header row in green
\toprule
\textbf{Model} & \textbf{Training RMSE} & \textbf{Training MAE} & \textbf{Testing RMSE} & \textbf{Testing MAE} \\
\midrule
QRNN          &  0.010322  & 0.007895  & {0.029297}  & {0.020059}  \\
QLSTM \cite{li2024quantum}        & 0.037125   &  0.030868  &  0.148198   &    0.132078 \\
QGRU \cite{li2024quantum}         &  0.014531  & 0.012054  &  0.136623  &  0.115907  \\
QTransformer \cite{zhang2025quantum}  &  0.029244  & 0.027604  & 0.243012   &   0.20540  \\
CRNN          &  0.010275  &  0.007784   & 0.086217  &  0.070392 \\
LSTM          & 0.089625  & 0.073255  & 0.214902  &   0.197111 \\
GRU           & 0.018507  & 0.015374  &   0.185229 &  0.163096 \\
Transformer   &   0.007895 &  0.005053 &  0.229551  &   0.201858 \\
\bottomrule
\end{tabular}
\caption{Prediction errors between QRNN, QLSTM, QGRU, QTransformer, CRNN, LSTM, GRU, and Transformer with multi-qubit system.}
\label{tab:1}
\end{table*}

\begin{table*}[ht]

\centering
\renewcommand{\arraystretch}{5}
\rowcolors{2}{blue!10}{blue!10} % Apply light blue to data rows
    \renewcommand{\arraystretch}{1.2} % Adjust row height

% \begin{adjustbox}{max width=\textwidth}
\begin{tabular}{|p{3cm}|p{1.5cm}|p{1.5cm}|p{1.8cm}|p{1.6cm}|}
% \begin{tabular}{lrrrrr}
\rowcolor{green!30} % Header row in green
\toprule
{Stock} & {Phase} & {Metric} & {CRNN} & {QRNN}   \\
\midrule
APPLE      & Test & RMSE &  0.105341   &  {0.029297}    \\
           &       & MAE  & 0.081971  &    {0.020059}\\
\midrule           
 SONY      
           & Test  & RMSE & 0.110905 & {0.036476 }\\
  
        &       & MAE & 0.100325 & {  0.026508 }\\
\midrule
TESLA     
           
           & Test  & RMSE&0.083813  & { 0.035643 }\\
           &       & MAE & 0.041538  & { 0.023245}\\
\midrule
NVIDIA   & Test  & RMSE& 0.206544 & {0.043539 }  \\
           &       & MAE &  0.177895 & { 0.035297 }  \\
\midrule
INTEL      
           & Test  & RMSE&0.122567     & {0.039529 }  \\
           &       & MAE &  0.102202  & {0.029904 }  \\
\midrule
GENERAL MOTOR   & Test  & RMSE& 0.050316  & { 0.013745}  \\
           &       & MAE & 0.038451&  {0.010920 } \\
\midrule
FORD    & Test  & RMSE& 0.039106  & {0.028047 }  \\
           &       & MAE  & 0.030748 & {0.021782 }  \\
\midrule

AAL     
           & Test  & RMSE& 0.052535  & {0.034016}  \\
           &       & MAE &0.038934 & {   0.026449}  \\
\midrule
 AMD    
           & Test  & RMSE& 0.102331  & { 0.037626   }\\
           &       & MAE &  0.069866& { 0.026427  }  \\
\midrule
GOOGLE     
           & Test  & RMSE& 0.110905   & { 0.036476 }\\
           &       & MAE &0.092325& {0.026508}  \\
\midrule
AMAZON    
           & Test  & RMSE& 0.123007     & { 0.029657  }\\
           &       & MAE & 0.104958& {0.021630}  \\
\bottomrule
\end{tabular}
\caption{Performance comparison of the proposed method compared to the baseline model with a multi-qubit system.}
\label{tab:2}
\end{table*}

\subsection{Convergence Behavior}

\begin{figure*}[!ht]
    \centering
    % --------- Second figure ---------
    \includegraphics[width=14cm]{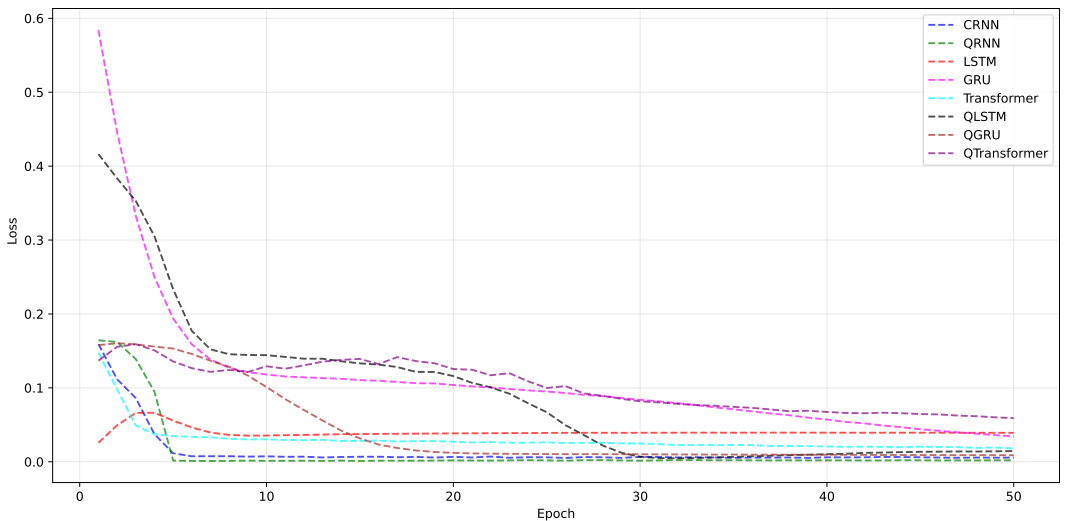}
    \caption{Comparison of losses between CRNN, QRNN, LSTM, GRU, Transformer, QLSTM, QGRU, and QTransformer.}
   \label{fig:5}
\end{figure*}

\begin{figure*}[htbp]
    \centering
    % Row 1
    \begin{minipage}[b]{0.45\textwidth}
        \centering
        \includegraphics[width=.96\linewidth]{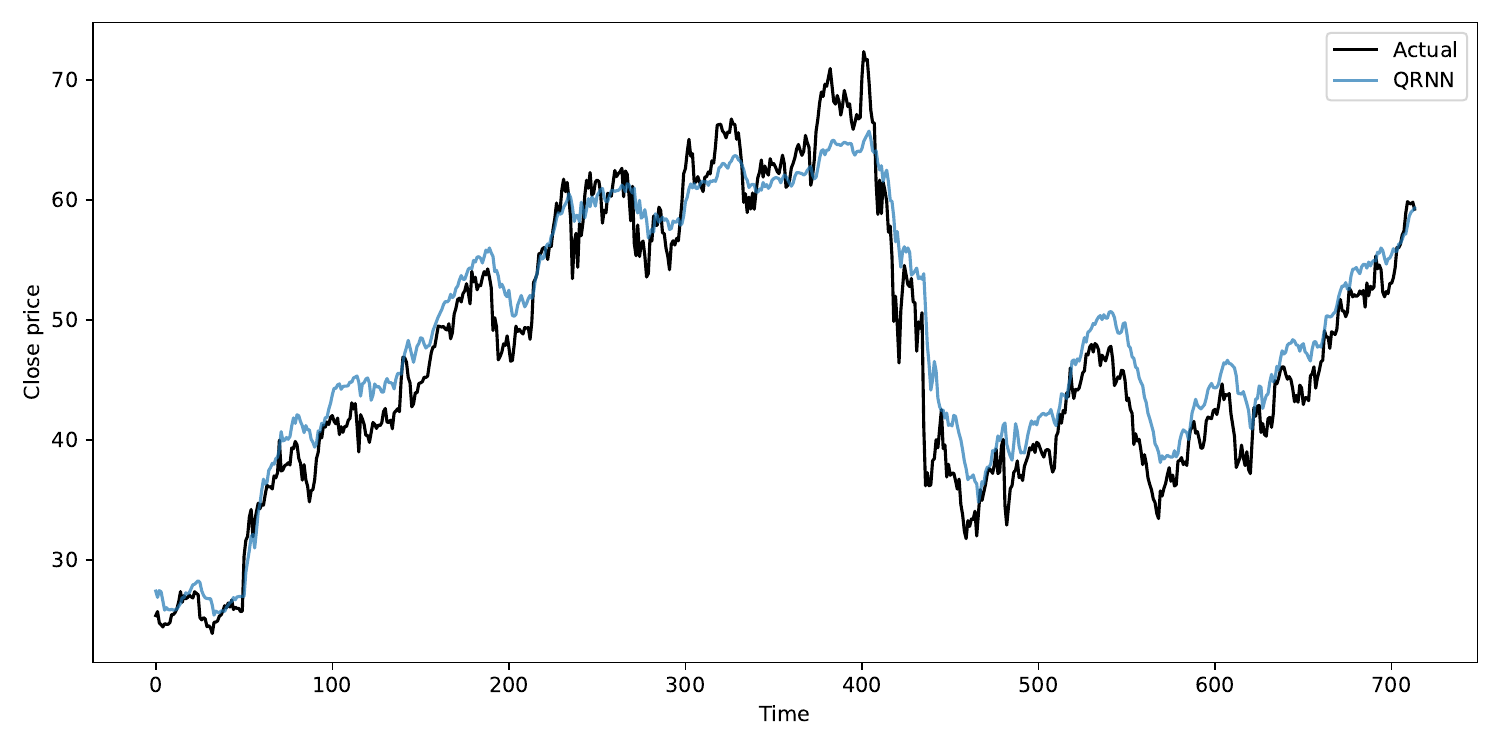}
        \subcaption{NVIDIA}
    \end{minipage}
    \hfill
    \begin{minipage}[b]{0.45\textwidth}
        \centering
        \includegraphics[width=.96\linewidth]{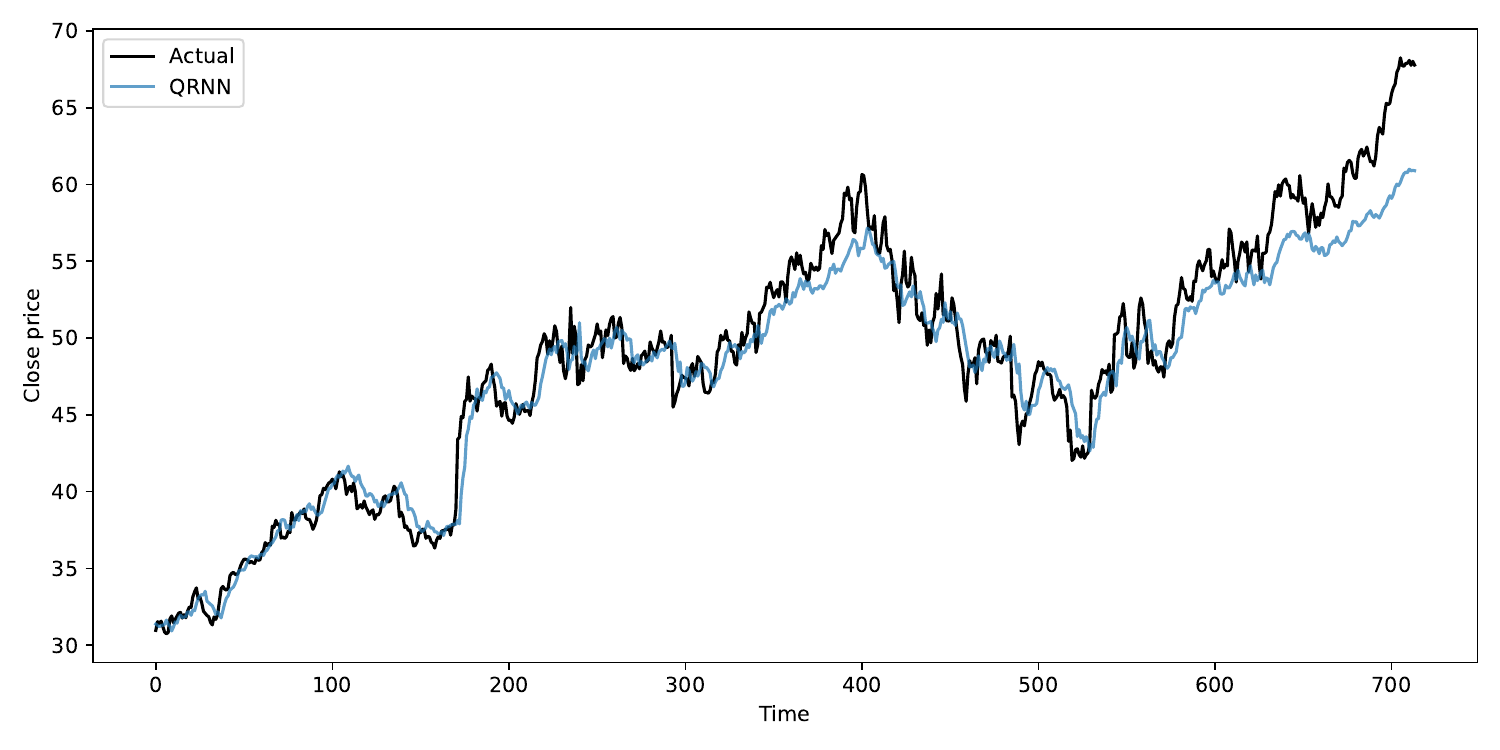}
        \subcaption{SONY}
    \end{minipage}
    
    \vspace{0.5cm}
    
    % Row 2
    \begin{minipage}[b]{0.45\textwidth}
        \centering
        \includegraphics[width=.96\linewidth]{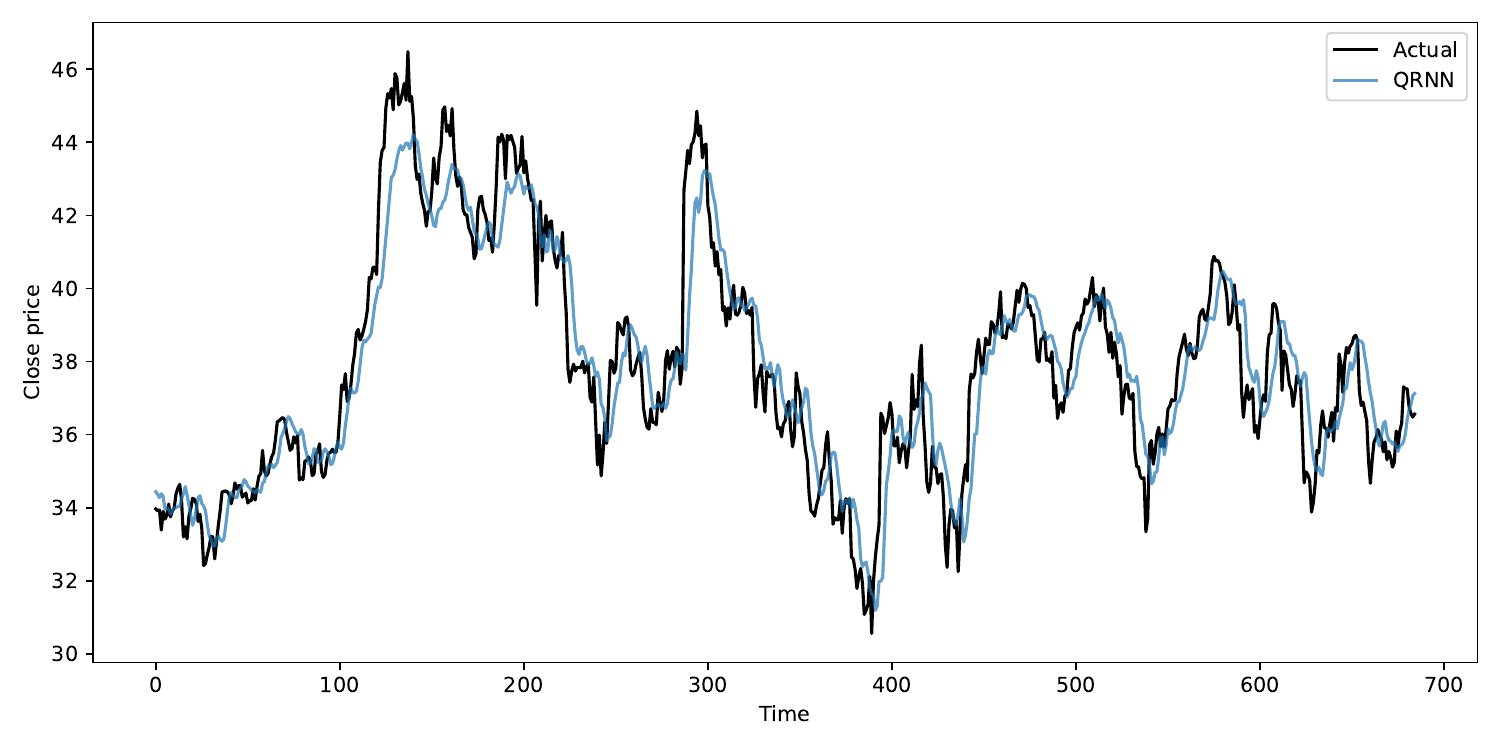}
        \subcaption{GENERAL MOTORS}
    \end{minipage}
    \hfill
    \begin{minipage}[b]{0.45\textwidth}
        \centering
        \includegraphics[width=.96\linewidth]{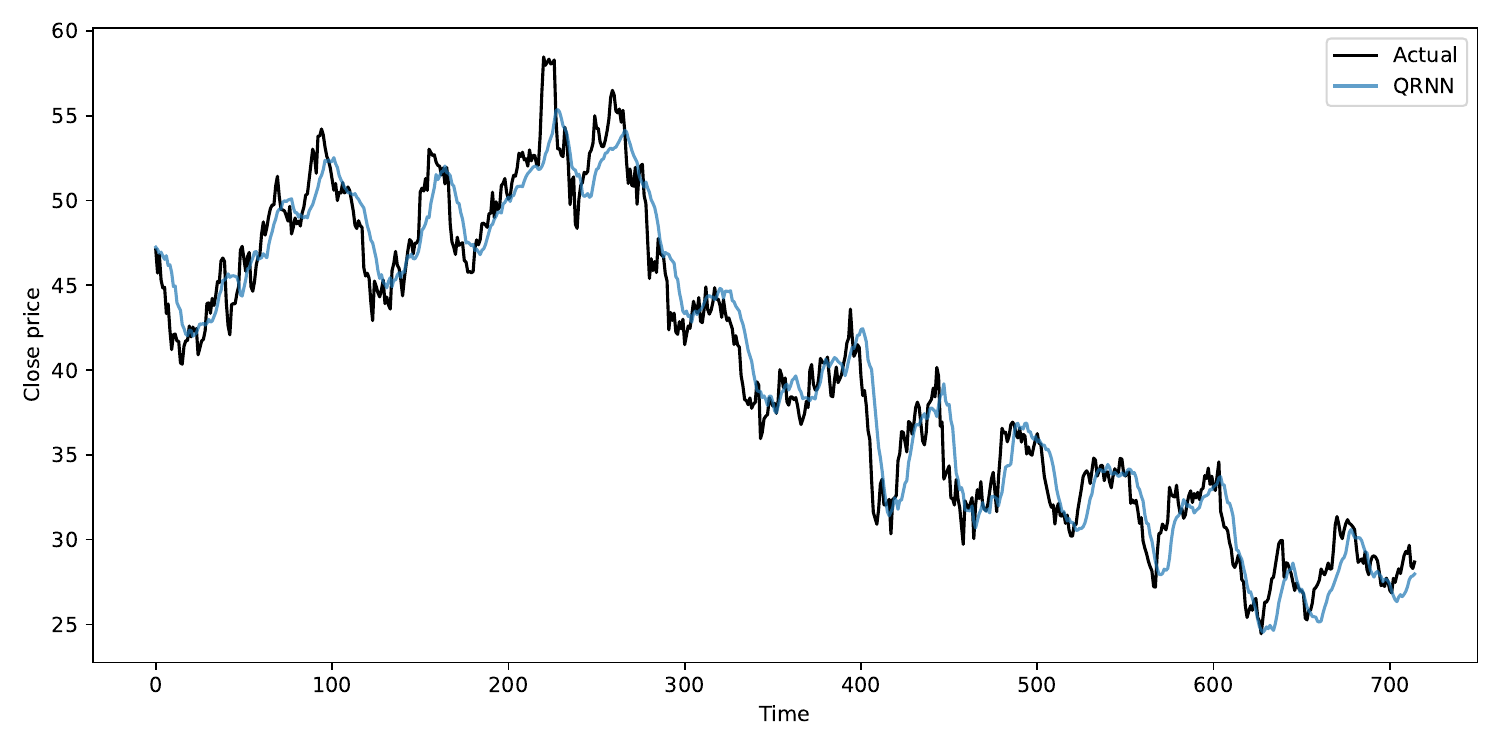}
        \subcaption{AAL}
    \end{minipage}

     % Row 3
    \begin{minipage}[b]{0.45\textwidth}
        \centering
        \includegraphics[width=.96\linewidth]{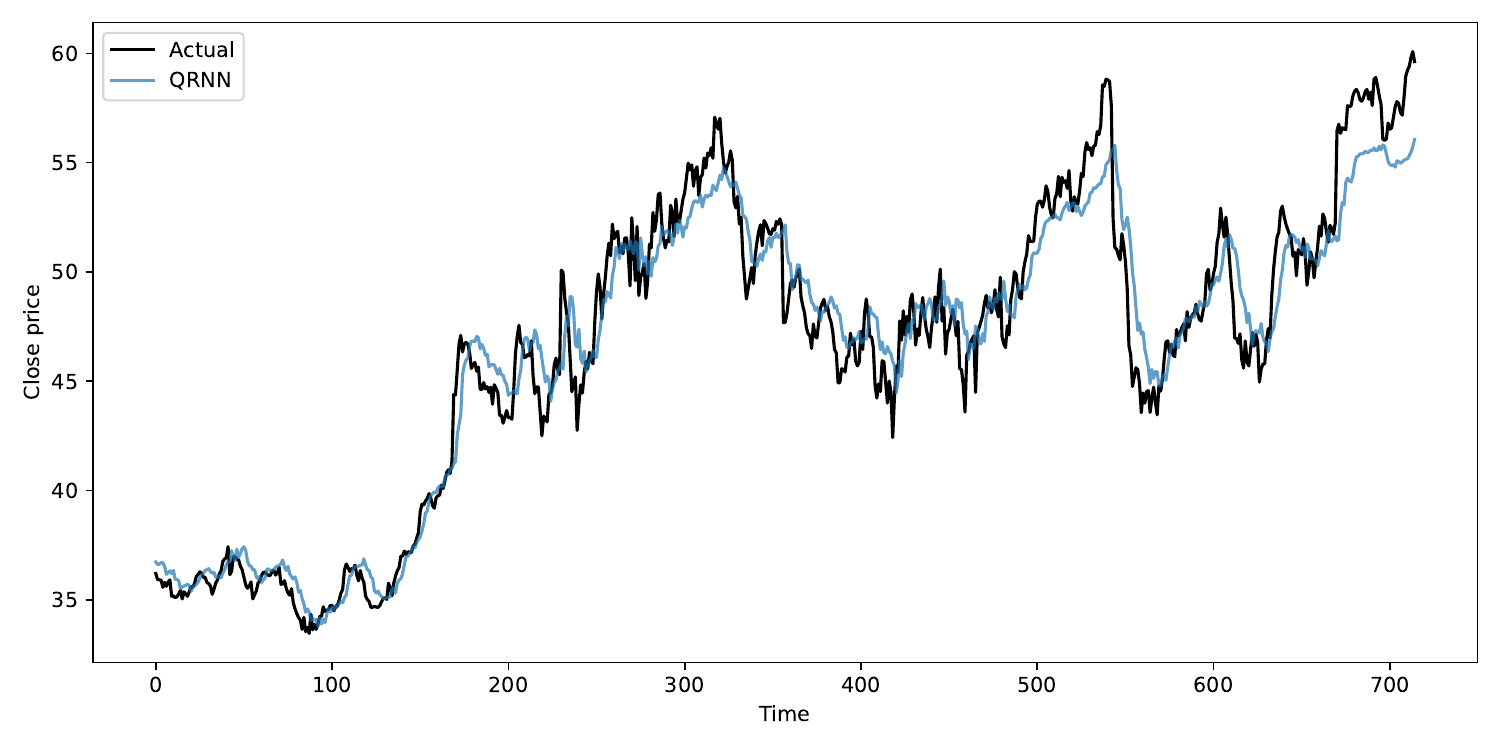}
        \subcaption{Intel}
    \end{minipage}
    \hfill
    \begin{minipage}[b]{0.45\textwidth}
        \centering
        \includegraphics[width=.96\linewidth]{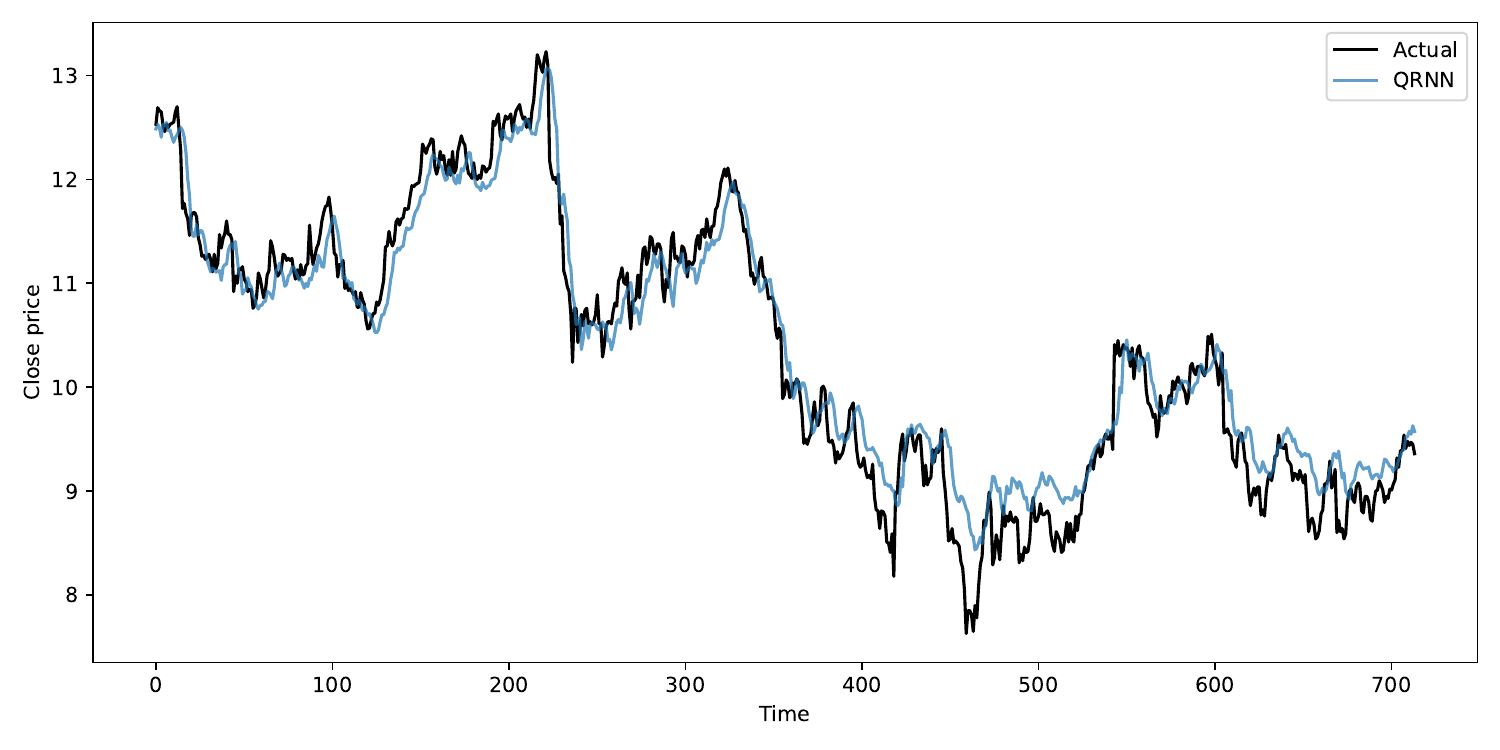}
        \subcaption{FORD}
    \end{minipage}

      % Row 4
    \begin{minipage}[b]{0.45\textwidth}
        \centering
        \includegraphics[width=.96\linewidth]{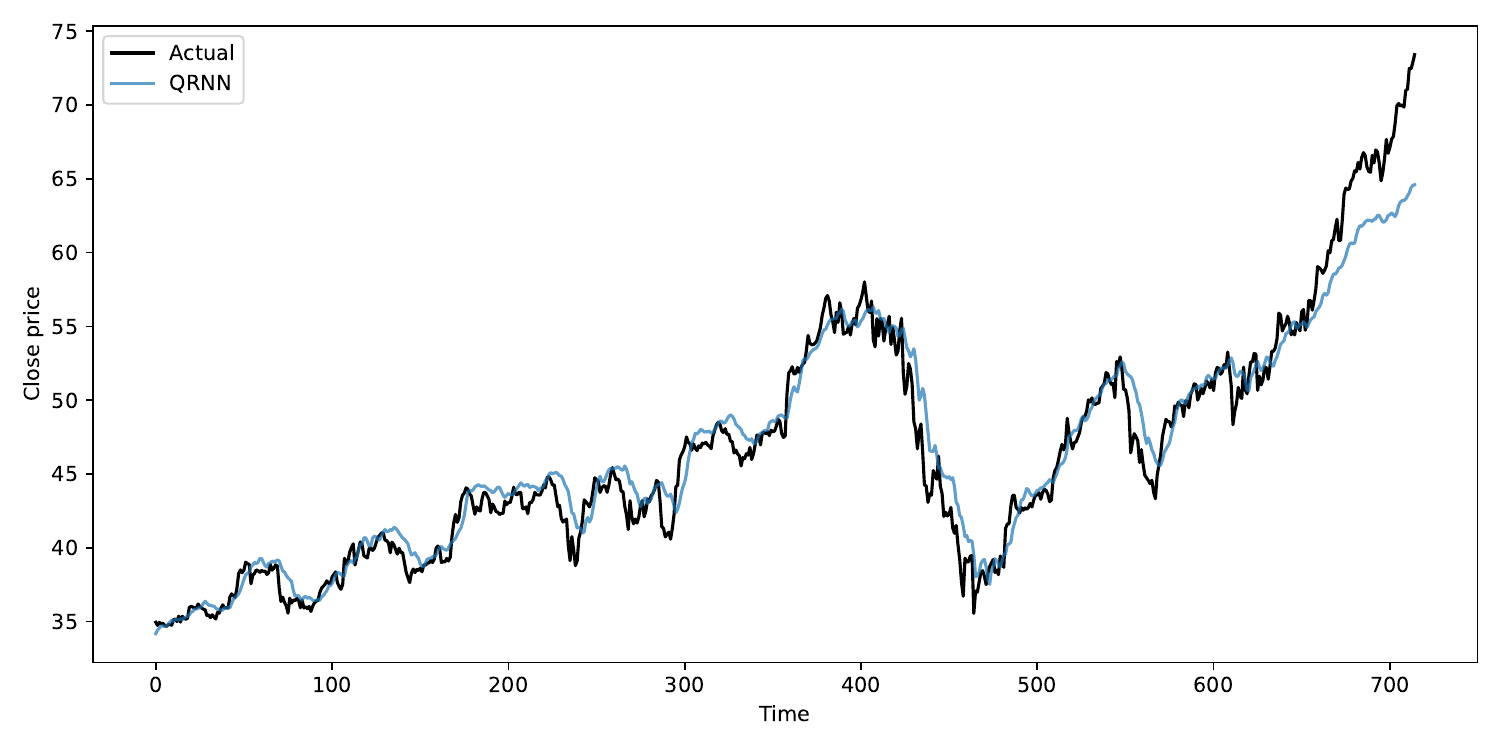}
        \subcaption{AAPL}
    \end{minipage}
    \hfill
    \begin{minipage}[b]{0.45\textwidth}
        \centering
        \includegraphics[width=.96\linewidth]{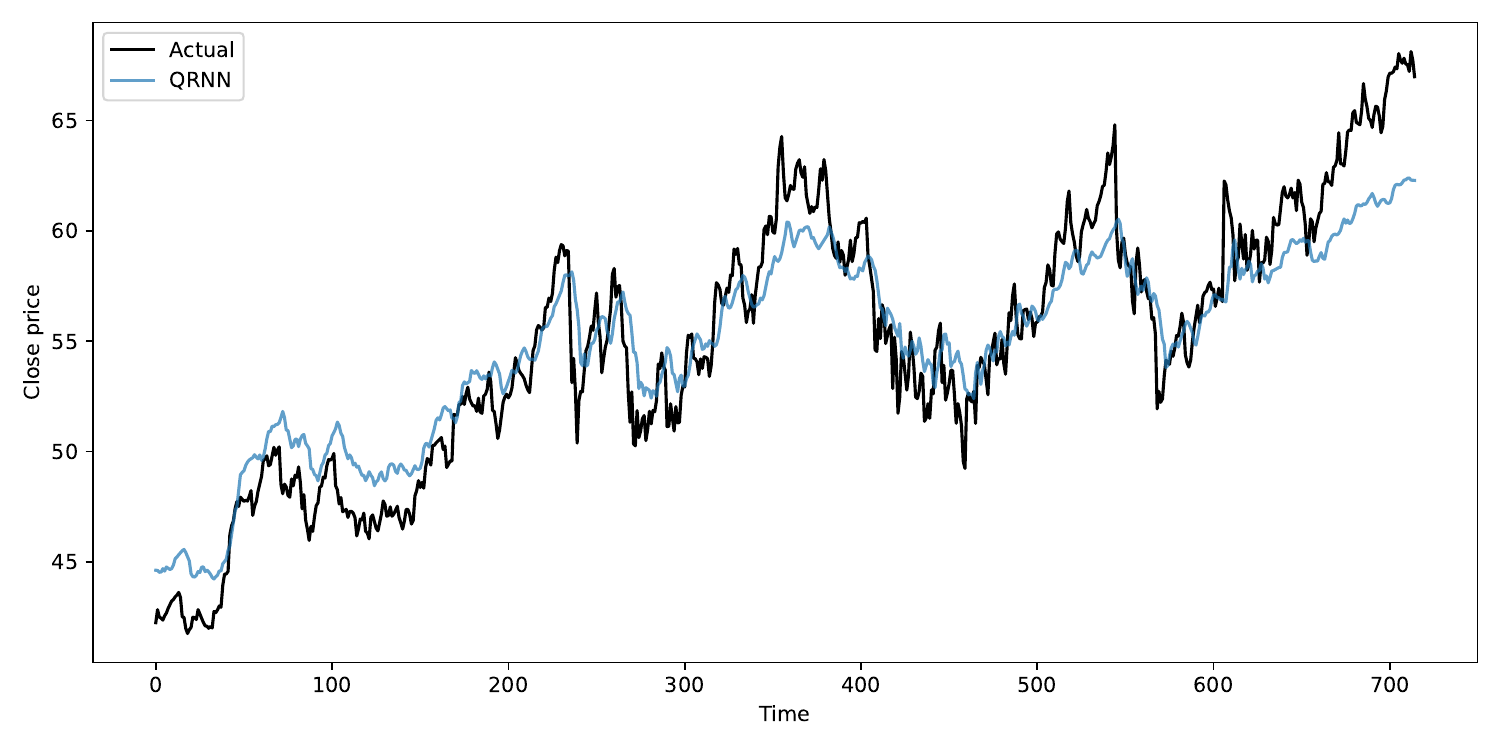}
        \subcaption{GOOGLE}
    \end{minipage}
    \hfill
    % Row 2
    \begin{minipage}[b]{0.45\textwidth}
        \centering
        \includegraphics[width=.96\linewidth]{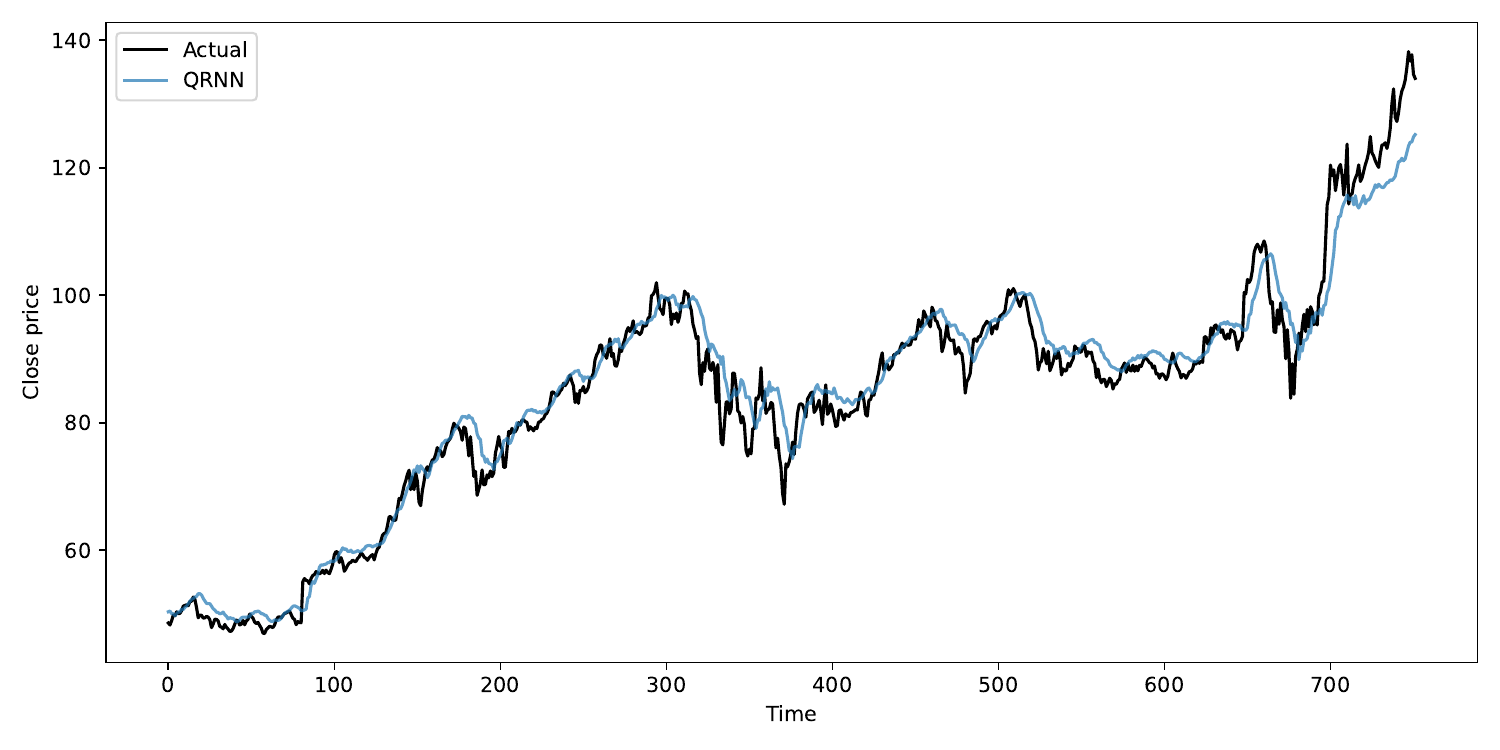}
        \subcaption{AMZN}
    \end{minipage}
    \hfill
    \begin{minipage}[b]{0.45\textwidth}
        \centering
        \includegraphics[width=.96\linewidth]{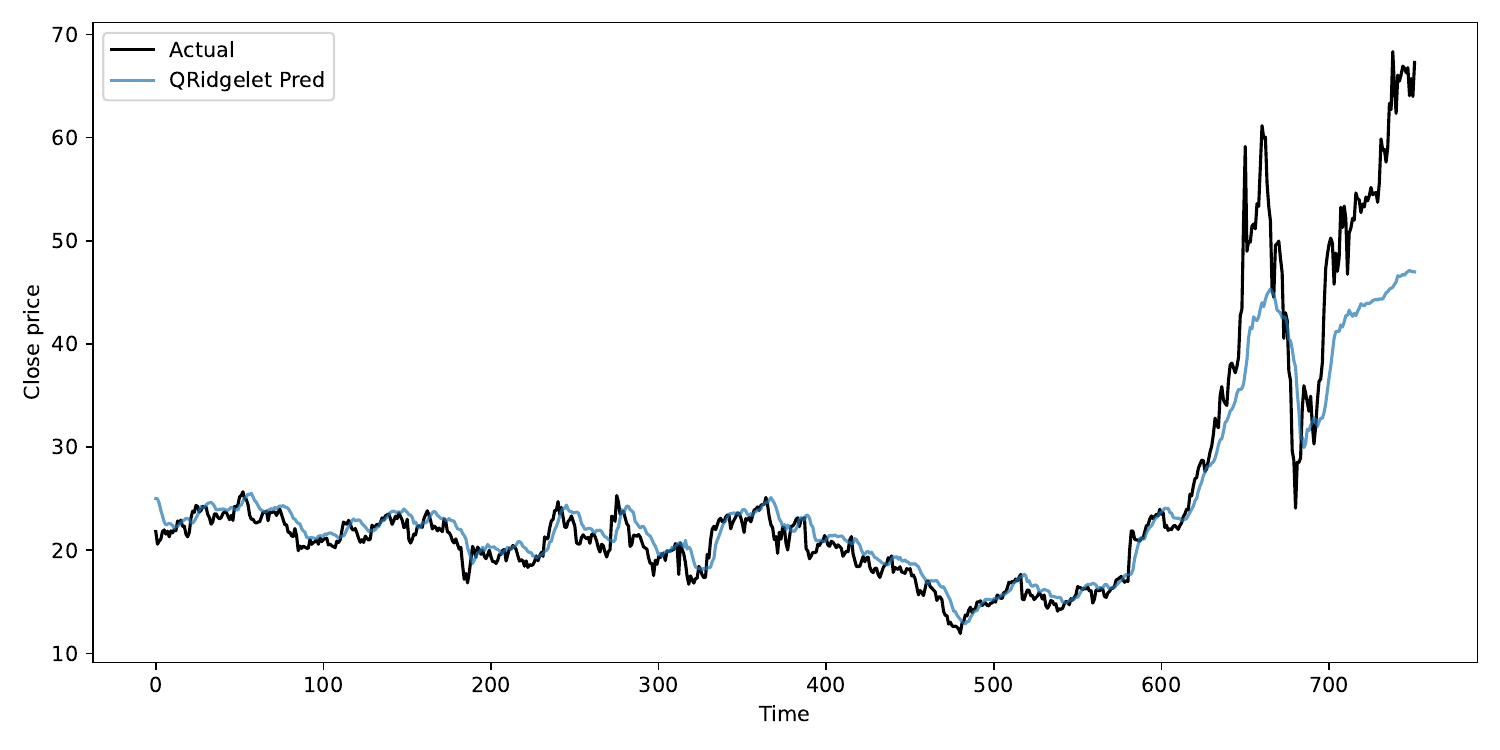}
        \subcaption{TSLA}
    \end{minipage}
    \caption{Comparison of the test data set of multiple stocks to compare predicted QRNN and real prices.}
    \label{fig:8}
\end{figure*}

Here, we begin by investigating how our proposed QRNN can improve accuracy and reduce training loss. We compare the performance of existing baseline quantum and classical models in detail to show that our proposed QRNN model performs better. Training losses, as shown in Fig. \ref{fig:5}, reveal that QRNN's training losses are consistently smaller and exhibit less variation than other quantum-classical models, including QLSTM, QGRU, QTransformer, CRNN, LSTM, GRU, and Transformer over different epochs. The precision of data depiction improves through the use of quantum encoding and the incorporation of classical data in a higher-dimensional Hilbert space. The convergence behavior of our proposed QRNN model indicates that it outperforms compared to quantum and classical models, such as QLSTM, QGRU, QTransformer, CRNN, LSTM, GRU, and Transformer for stock price prediction. The QRNN model exhibits a steady and uniform decrease in training loss over epochs, as shown in Fig. \ref{fig:5}, indicating stable and effective learning. Although CRNN performs competitively and comes second in terms of convergence rates, QRNN outperforms it because it maintains lower loss values throughout the training process. However, the QTransformer, QLSTM, QGRU, Transformer, GRU, and LSTM models show larger final loss and slower convergence, indicating a decreased capacity to adjust to the intricate and erratic patterns found in stock price prediction, as shown in Fig. \ref{fig:5}. By monitoring performance over 50 epochs, it is clear that QRNN can capture both short-term volatility and long-term dependency on stock prices. It also converges more quickly and reaches a lower error plateau. These outcomes demonstrate QRNN's resilience and dependability in predicting stock prices in erratic market situations, making it an innovative approach for stock price prediction. Furthermore, Fig. \ref{fig:8} displays the testing performance of our proposed QRNN model with actual stock data. Our model's capacity to capture both long-term trends and short-term changes is demonstrated by how closely the predicted prices match the actual prices throughout the test period. These results demonstrate a good generalization ability and robustness on unseen data, as indicated by the strong correlation between predicted and actual values, as shown in Fig. \ref{fig:8}.

\subsection{Portfolio Performance Evaluation}

In this study, we have considered 12 stocks from the various financial sectors to construct an optimal portfolio. The selected stocks include:  AAPL, MSFT, AMZN, NVDA, GOOGL, TSLA, AAL, SONY, INTC, AMD, F, and GM. The objective of our framework is to construct an optimal portfolio of fixed cardinality K=5. Here, the portfolio size of 5 is selected to ensure sufficient diversification and computational tractability while preserving a tractable combinatorial search space. Our configuration offers a practical and computationally feasible framework for portfolio optimization based on QRNN-QAOA. Here, we evaluate the comparative analysis of our model against the baseline portfolio construction strategies, including Top-K, equal-weight, and random selection, as presented in  Table \ref{tab:QAOA_result}. From Table \ref{tab:QAOA_result}, we can observe that our model achieves a better Sharpe ratio, indicating a superior risk-adjusted performance compared to other approaches. Whereas the Top-K approach gives a slightly higher predicted return, its actual return is significantly lower, reflecting its inability to account for risk and asset correlations, because it fails to translate them into actual gains due to a lack of risk consideration. From Table \ref{tab:QAOA_result}, it also shows that our model maintains a closer alignment between predicted and actual returns, suggesting more reliable and stable portfolio selection. Furthermore, the variance of our approach is significantly lower than that of Top-K, equal-weight, and random portfolios, demonstrating effective risk control through optimization. However, the equal-weight and random strategies exhibit both lower returns and higher risk, confirming their limited effectiveness compared to model-driven approaches. This finding shows that our method provides more stable and reliable portfolio returns.
\begin{table*}[ht]
\centering
\renewcommand{\arraystretch}{5}
\rowcolors{2}{blue!10}{blue!10} % Apply light blue to data rows
    \renewcommand{\arraystretch}{1.2} % Adjust row height

\begin{tabular}{|p{3.5cm}|p{2.3cm}|p{2cm}|p{2.5cm}|p{1.8cm}|}
\rowcolor{green!30} % Header row in green
\toprule
\textbf{Method} & \textbf{Predicted Return} & \textbf{Actual Return} & \textbf{Variance} & \textbf{Sharpe Ratio} \\
\midrule
Our Approach          &  0.0064  &  0.0049  & {0.00020161}  & { 0.9281}  \\
Top-K       &  0.0068 &   0.0038 &  0.00039821    &  0.6311  \\
Equal Weight \cite{demiguel2009optimal}        &   0.0052 &  0.0029  &  0.00055620 &  0.4632 \\
Random  & 0.0046  &  0.0022  & 0.00069512 &  0.1293 \\

\bottomrule
\end{tabular}
\caption{Portfolio performance comparison across the different investment strategies.}
\label{tab:QAOA_result}
\end{table*}

\subsection{Limitation}
In this study, our model is constructed using logical qubits as the primary computational units, where each logical qubit is encoded using a 3-qubit repetition QEC technique.  This leads to an expansion to physical qubits, where $\text{Physical qubits} = 3 \times \text{Logical qubits}$ represents the total number of physical qubits. The density matrix simulation is also required to incorporate a noise model that includes bit-flip, phase-flip, and depolarizing channels. This distinction is critical because pure state simulation scales as $2^n$, whereas mixed state simulation scales as $4^n$ due to the storage of a full density matrix. Consequently, memory requirements grow exponentially with the number of physical qubits. Specifically, for 1 logical qubit, we need 3 physical qubits, the state size is $4^3 = 64$, which is very small and easily manageable; for 2 logical qubits, it requires 6 physical qubits, the state size becomes $4^6 = 4096$, which remains tractable; and for 3 logical qubits, it needs 9 physical qubits, the state size increases to $4^9 = 262{,}144$, which is large but still computationally feasible. However, when scaling to 4 logical qubits, which require 12 physical qubits, the state size grows to $4^{12} = 16{,}777{,}216$, which is prohibitively large and leads to GPU memory exhaustion. This results in a GPU memory consumption approaching $\sim 95$GB, which ultimately causes a CUDA out-of-memory (OOM) error. As we increase the number of qubits, computationally, it is difficult to tackle. Notably, our model limitation is not due to implementation inefficiency, but is a fundamental consequence of the exponential complexity of quantum simulation. Large-scale quantum models are much more difficult to train, even if there is enough memory. Increasing the number of qubits creates significant optimization problems, such as vanishing gradients, higher levels of noise accumulation, and more challenging training dynamics.

\section{Conclusion}\label{section:8}
In this study, we presented a quantum-classical framework for financial forecasting and portfolio optimization by combining QRNN with QAOA-based portfolio selection.  QRNN model that incorporates PQCs and ridgelet-based feature transformations. Our model effectively captures high-dimensional features and directional patterns in the data by incorporating ridgelet characteristics and estimating the expected return for multiple assets. The predicted return is used to guide portfolio construction within constrained optimization settings, where QAOA is used to identify an efficient asset subset under risk-return trade-off consideration. We established a theoretically grounded extension to multi-qubit systems, in which the ridgelet representation is combined with quantum circuit dynamics to improve expressive power, derived from a single-qubit formulation. This integration enables the model to improve representation learning by utilizing both quantum entanglement and systematic feature extraction. Our work focuses on analyzing the model in simulated quantum settings. We investigated the behavior of ridgelet-enhanced quantum frameworks for the presence of deficiencies by adding noise channels and examining both full and partial QEC approaches. Our results demonstrated that noise reduces performance and the incorporation of error-correction techniques significantly enhances stability, with partial QEC providing a useful compromise between computational complexity and robustness.  However, the capacity of our proposed framework is proven by experimental findings in several qubit configurations. Additionally, we further validated the actual financial data that demonstrates the significance of our model, which has superior predictive performance compared to other quantum and traditional deep learning techniques. Furthermore, we demonstrate the practicality of our approach for detecting high-potential stocks for portfolio development using a QAOA-based portfolio optimization approach. Despite encouraging outcomes, our proposed QRNN model is computationally challenging due to repeated quantum circuit evaluations.  It creates opportunities for further study on the integration of scalable quantum structures with sophisticated functional representations.

\section*{Acknowledgements}
Sanjay Kumar Mohanty acknowledges partial support from the Department of Science and Technology, Government of India, under grant SR/FST/MS-II/2023/139(C) at VIT Vellore.

\bibliography{References}
\end{document}